\documentclass{article} 
\usepackage{iclr2027_conference,times}

\usepackage{amsmath,amsfonts,bm}

\def\eqref#1{equation~\ref{#1}}

\def\1{\bm{1}}

\DeclareMathAlphabet{\mathsfit}{\encodingdefault}{\sfdefault}{m}{sl}
\SetMathAlphabet{\mathsfit}{bold}{\encodingdefault}{\sfdefault}{bx}{n}

\usepackage{amsmath}
\usepackage{amssymb}
\usepackage{algorithm}
\usepackage{algorithmic}
\usepackage{hyperref}
\usepackage{url}
\hypersetup{colorlinks, breaklinks, citecolor=[rgb]{0, 0.44, 0.737},
anchorcolor=[rgb]{1, 0.44, 0.737}, linkcolor=[rgb]{0, 0.44, 0.737},
urlcolor=[rgb]{0, 0.44, 0.737} 
}
\usepackage{wrapfig}
\usepackage{graphicx}
\usepackage{booktabs}
\usepackage{makecell}
\usepackage{multirow}
\usepackage[table]{xcolor}
\usepackage{arydshln}
\usepackage{amsthm}
\usepackage{pifont}
\usepackage{caption}
\usepackage{xspace}

\usepackage{listings}
\usepackage{xcolor}
\usepackage{float}
\usepackage{placeins}
\definecolor{link}{RGB}{214,51,132}

\newtheorem{proposition}{Proposition}

\newcommand{\cmark}{\textcolor{green!60!black}{\ding{51}}}
\newcommand{\xmark}{\textcolor{red}{\ding{55}}}
\newcommand{\soloq}{\textsc{SoloQ}\xspace}

\title{SoloQ: Calibration-Free Quantization for Diffusion Language Models}

\author{
Donghyun Lee$^1$, Arkapravo Ghosh$^1$, Varun Manjunath$^1$, Bumjoon Kyle Rhee$^1$,\\\textbf{ Hyunho Kook$^1$, Shiting Xiao$^2$, Youngeun Kim$^3$, Priyadarshini Panda$^1$} \\
$^1$University of Southern California, $^2$Yale University, $^3$Korea University \\
\texttt{ donghyun.lee.1@usc.edu}
}
\iclrfinalcopy 
\begin{document}

\maketitle
\fancyhead{}

\vspace{-4mm}
\begin{abstract}
\vspace{-2mm}
Diffusion large language models (dLLMs) have emerged as a promising alternative to autoregressive language models through bidirectional diffusion-based token generation. However, their growing model sizes and high inference costs make efficient deployment challenging: full-sequence denoising repeatedly invokes compute-intensive forward passes, while block-diffusion models additionally introduce a memory-intensive KV-cache. Low-bit weight--activation quantization is therefore attractive, yet existing dLLM post-training quantization methods rely on calibration data despite activation distributions shifting across masking states and denoising steps. We present \textbf{\soloq}, a calibration-free quantization framework that maps weights and activations into a normalized rotated basis with a predictable marginal distribution, enabling data-independent quantization. \soloq combines a structured K-RPBH rotation with a lightweight rescaling correction for calibration-free quantization. Its predictable post-rotation distribution supports both distribution-matched codebooks and hardware-native NVFP4.
For block-diffusion models, \soloq further applies commit-time KV-cache quantization to compress persistent states without perturbing the actively denoised block. Across full-sequence dLLMs (LLaDA and Dream) and block-diffusion dLLMs
(Fast-dLLM v2 and Nemotron-Labs-Diffusion), \soloq retains accuracy under 4-bit quantization and outperforms calibration-based baselines on knowledge- and reasoning-intensive benchmarks. With NVFP4, \soloq reduces peak memory by up to $2.61\times$ and accelerates end-to-end inference by up to $2.24\times$. The code is available at \href{https://github.com/Intelligent-Computing-Lab-Panda/SoloQ}{\small\texttt{https://github.com/Intelligent-Computing-Lab-Panda/SoloQ}}.
\end{abstract}

\vspace{-4mm}
\section{Introduction}
\vspace{-2mm}
Large language models (LLMs) have achieved remarkable progress across a wide range of language understanding, reasoning, and generation tasks~\citep{brown2020language, achiam2023gpt, touvron2023llama, team2023gemini}, with autoregressive (AR) modeling remaining the dominant paradigm for text generation. 
In parallel, diffusion models~\citep{bao2023all, peebles2023scalable} have become a powerful generative framework for continuous domains such as image and video generation~\citep{esser2024scaling, wu2025qwen, kong2024hunyuanvideo, yang2025cogvideox}, where iterative denoising enables high-quality and flexible generation. 
Motivated by these advances, recent diffusion large language models (dLLMs) extend diffusion-based generation to discrete text, replacing strictly left-to-right decoding with iterative masked denoising and bidirectional context. Representative models such as LLaDA~\citep{nie2026large} and Dream~\citep{ye2025dream} have demonstrated competitive language understanding, reasoning, and generation capabilities, suggesting that strong language modeling is not inherently tied to AR decoding. More recently, block-diffusion models~\citep{wu2026fast, fu2026nemotron, team2026diffusiongemma} further bridge diffusion-based generation and practical inference by generating multiple tokens within a block and reusing previously generated context through caching. Together, these developments establish diffusion-based language modeling as a promising alternative to conventional AR generation.

\begin{figure*}[t]
    \centering
    \includegraphics[
        width=\textwidth
    ]{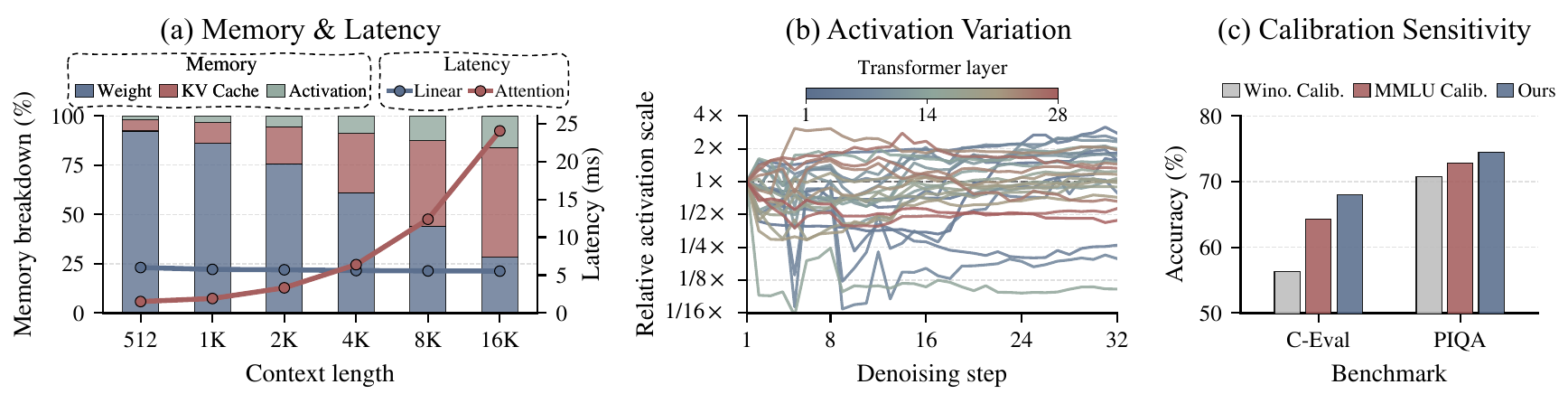}
    \vspace{-7mm}
   \caption{
Profiling and motivation for \soloq.
(a) Memory breakdown at batch size 128 and aggregate linear/attention latency at batch size 1 for Fast-dLLM v2~\citep{wu2026fast} across context lengths on an H200 GPU.
(b) Layer-wise variation in \texttt{down\_proj} input activation scale across denoising steps, normalized to the first step.
(c) Accuracy of PTQ~\citep{ashkboos2024quarot,frantar2022gptq} with WinoGrande/MMLU calibration versus calibration-free \soloq.
}
    \label{fig:motivation}
    \vspace{-5.1mm}
\end{figure*}

Despite this progress, efficient dLLM inference remains challenging as model size and context length grow.
Unlike AR decoding, dLLMs iteratively refine multiple tokens in parallel across denoising steps, while long-context inference increases the memory and computation associated with activations and KV-cache.
To better understand these bottlenecks, we profile Fast-dLLM v2~\citep{wu2026fast} from memory and latency perspectives, as shown in Figure~\ref{fig:motivation}(a).
At short contexts, weights dominate the memory footprint, but the relative contribution of activations and the KV-cache grows with context length.
In terms of latency, attention  becomes expensive as the sequence length grows.
In dLLMs, these costs are repeatedly incurred across denoising steps, making weight-only compression insufficient for efficient inference.
Instead, low-bit quantization of weights, activations, and KV-cache can reduce both computation and memory across context regimes.

Post-training quantization (PTQ) has been widely studied for AR LLMs to reduce memory and computation without retraining.
Existing LLM PTQ methods span several complementary paradigms, including second-order methods that exploit Hessian information for low-bit weight quantization~\citep{frantar2022gptq,li2025gptaq,lee2026kronq}, scaling- and salience-based methods that mitigate activation outliers through scaling or reparameterization~\citep{xiao2023smoothquant,wei2023outlier,shao2024omniquant}, and rotation-based methods that transform weights and activations into more quantization-friendly representations~\citep{chee2023quip,ashkboos2024quarot,liu2025spinquant,hu2025ostquant}.
These methods enable efficient low-bit inference while preserving accuracy.


However, quantization on dLLMs is more challenging than AR LLMs. During diffusion decoding, the model repeatedly encounters different masking states and denoising steps, leading to substantial variation in activation distributions and quantization sensitivity throughout generation, as shown in Figure~\ref{fig:motivation}(b).
Existing dLLM PTQ methods~\citep{zhang2026quant,xu2025dllmquant,yan2026star,huang2026fair} rely on calibration data to capture or compensate for these state- and time-dependent variations.
Figure~\ref{fig:motivation}(c) illustrates this sensitivity for a representative calibration-based PTQ method~\citep{ashkboos2024quarot,frantar2022gptq}: changing calibration data from WinoGrande to MMLU alters accuracy across the benchmarks, showing that calibration choices may not generalize across tasks.
This motivates a different direction: \textit{rather than tracking a changing distribution, can we transform it into a predictable form that can be quantized without calibration?}

To realize this principle, we propose \textbf{\soloq}, a calibration-free dLLM quantization framework that avoids calibration-set dependence while maintaining strong performance, as shown in Figure~\ref{fig:motivation}(c).
Rather than adapting to activation distributions that vary across masking states and denoising steps, \soloq transforms weights and activations into a normalized rotated space with a predictable dimension-dependent marginal distribution.
This enables data-independent quantization that can be shared across diffusion states without calibration.
To support arbitrary transformer dimensions, \soloq introduces a Kronecker-extended randomized permuted block-Hadamard (K-RPBH) rotation that combines efficient local mixing with global cross-block mixing.
We further introduce a scale-only bias correction that restores the radial component of low-bit quantized representations with negligible overhead.
For block-diffusion models, \soloq extends the same calibration-free design to persistent KV states through commit-time KV-cache quantization.
Built on these components, \soloq supports both distribution-matched codebook quantization and hardware-native NVFP4 without chasing activation distributions. Our contributions are summarized as follows:
\vspace{-1.5mm}
\begin{itemize}
    \item We introduce \textbf{\soloq}, a calibration-free dLLM quantization framework combining K-RPBH rotation with a lightweight scale-only bias correction.
    \item The resulting single predictable distribution supports both distribution-matched codebooks and hardware-native NVFP4 within the same calibration-free framework.
    \item We extend \soloq to block-diffusion models through commit-time KV-cache quantization, covering weights, activations, and persistent KV states.
    \item Across full-sequence and block-diffusion dLLMs, \soloq achieves strong 4-bit accuracy with up to a $2.61\times$ reduction in peak memory and $2.24\times$ faster end-to-end inference.
\end{itemize}


\section{Related Work}
\vspace{-3mm}
\subsection{Diffusion Large Language Models}
\vspace{-2mm}
Diffusion-based text generation has evolved from early discrete and masked diffusion models~\citep{austin2021structured,li2022diffusion,he2023diffusionbert} to large-scale diffusion language models that perform iterative denoising with bidirectional context.
Recent full-sequence models such as LLaDA~\citep{nie2026large}, LLaDA-1.5~\citep{zhu2026llada}, and Dream~\citep{ye2025dream} demonstrate competitive language understanding, reasoning, and generation at billion-parameter scale.
Beyond full-sequence diffusion, block-based and hybrid approaches improve decoding efficiency by generating multiple tokens per block and reusing previously generated context, including Block Diffusion~\citep{arriola2025block}, Fast-dLLM~\citep{wu2026fast1}, Fast-dLLM v2~\citep{wu2026fast}, and LLaDA2.X~\citep{bie2025llada20scalingdiffusionlanguage,bie2026llada21speedingtextdiffusion}.
More recent systems further broaden this design space: Nemotron-Labs-Diffusion~\citep{fu2026nemotron} unifies autoregressive, diffusion, and self-speculative decoding, Nemotron-TwoTower~\citep{reda2026nemotron} separates autoregressive context modeling from diffusion denoising, and DiffusionGemma~\citep{team2026diffusiongemma} scales block-diffusion generation to a 26B mixture-of-experts model.
These developments highlight the rapid progress of diffusion-based language modeling toward scalable and practical text generation.

\vspace{-2mm}
\subsection{Calibration-Free and Data-Independent Quantization}
\vspace{-2mm}
A complementary line of research avoids calibration altogether by deriving quantizers from analytically known or data-independent distributions.
TurboQuant~\citep{zandieh2026turboquant} randomly rotates input vectors to induce a known coordinate distribution, enabling coordinate-wise optimal quantization.
PolarQuant~\citep{han2025polarquant} similarly exploits an analytically known distribution after random preconditioning and polar transformation for online vector and KV-cache quantization.
OrbitQuant~\citep{lee2026orbitquant} extends this principle to image and video diffusion transformers using randomized permuted block-Hadamard rotations for calibration-free weight--activation quantization.
ConvRot~\citep{huang2025convrot} applies group-wise Hadamard rotations for W4A4 diffusion-transformer quantization, while DVD-Quant~\citep{li2026dvd} combines data-free grid refinement, rotation, and adaptive bit-width allocation for video diffusion transformers.
\soloq builds on this distributional perspective for diffusion language models, extending calibration-free quantization to arbitrary transformer dimensions and persistent KV states in block-diffusion models.

\vspace{-2mm}
\subsection{Post-Training Quantization for dLLMs}
\vspace{-2mm}
Recent work has begun to develop PTQ methods specifically for diffusion language models.
Quant-dLLM~\citep{zhang2026quant} introduces masked calibration simulation for timestep-dependent masking and targets extreme low-bit weight quantization.
DLLMQuant~\citep{xu2025dllmquant} accounts for temporal and masking variations for weight--activation quantization, while STaR-Quant~\citep{yan2026star} addresses state-dependent activation disparity and temporal error accumulation.
FAIR-Calib~\citep{huang2026fair} improves calibration by emphasizing fragile token-commitment states.
Despite their designs, existing dLLM PTQ methods remain calibration-based and focus on full-sequence diffusion models.
To our knowledge, \soloq is the first calibration-free framework to support KV-cache quantization for block-diffusion models while also covering full-sequence dLLMs.

\vspace{-2mm}
\section{Preliminary}
\vspace{-2mm}
\subsection{Diffusion Language Models}
\vspace{-2mm}
Let $x_0=(x_0^1,\ldots,x_0^L)\in\mathcal{V}^L$ denote a clean token sequence of length $L$.
At diffusion timestep $t\in[0,1]$, each token is independently preserved with probability $\alpha_t$ or replaced by a special mask token $\mathrm{M}$:
\begin{equation}
q_{t|0}(x_t\mid x_0)
=
\prod_{i=1}^{L}
\left[
\alpha_t \,\delta_{x_0^i}(x_t^i)
+
(1-\alpha_t)\,\delta_{\mathrm{M}}(x_t^i)
\right],
\label{eq:dllm_forward}
\end{equation}
where $x_t$ is the partially masked sequence and $\delta_a(\cdot)$ denotes a point mass at token $a$.
The denoising model predicts masked tokens from bidirectional context,
\begin{equation}
p_\theta(x_0^i\mid x_t),
\qquad i \ \text{s.t.}\ x_t^i=\mathrm{M},
\label{eq:dllm_denoise}
\end{equation}
and progressively unmasks the sequence over multiple steps.
Full-sequence dLLMs~\citep{nie2026large, ye2025dream} repeatedly process the entire sequence, whereas block-diffusion models~\citep{wu2026fast, fu2026nemotron} denoise one block at a time while reusing KV states from previously committed blocks.

\subsection{Predictable Marginals under Random Rotation}
\vspace{-2mm}
Consider a nonzero vector $x\in\mathbb{R}^{d}$ with normalized direction
$\tilde{x}=x/\|x\|_2$.
Under a Haar-random rotation $\mathbf{R}$, $z=\mathbf{R}\tilde{x}$ is uniformly distributed on the unit sphere, independent of the direction of $x$~\citep{zandieh2026turboquant}.
Consequently, each coordinate follows the fixed spherical marginal
\begin{equation}
f_d(u)
=
\frac{\Gamma(d/2)}
{\sqrt{\pi}\,\Gamma((d-1)/2)}
(1-u^2)^{(d-3)/2},
\quad
u\in[-1,1]
\approx
\mathcal{N}\!\left(0,\frac{1}{d}\right),
\label{eq:fixed_marginal}
\end{equation}
where the Gaussian approximation becomes increasingly accurate as $d$ grows.
Thus, random rotation maps arbitrary normalized representations to a predictable Gaussian-like distribution, enabling analytic quantizer design without calibration.
OrbitQuant~\citep{lee2026orbitquant} realizes this principle using randomized permuted block-Hadamard (RPBH) rotations.
For $d=kh$, with Hadamard block size $h$ and $k$ blocks, RPBH applies a global random permutation followed by independent randomized Hadamard transforms within each block, providing $O(d\log h)$ mixing without dense-rotation cost.
However, its blockwise structure can leave residual inter-block energy imbalance, which becomes more pronounced as the number of blocks grows due to limited cross-block mixing.


\vspace{-3mm}
\section{Method}
\vspace{-3mm}
\subsection{Overview}
\begin{figure}[t]
\centering
\includegraphics[width=\linewidth]{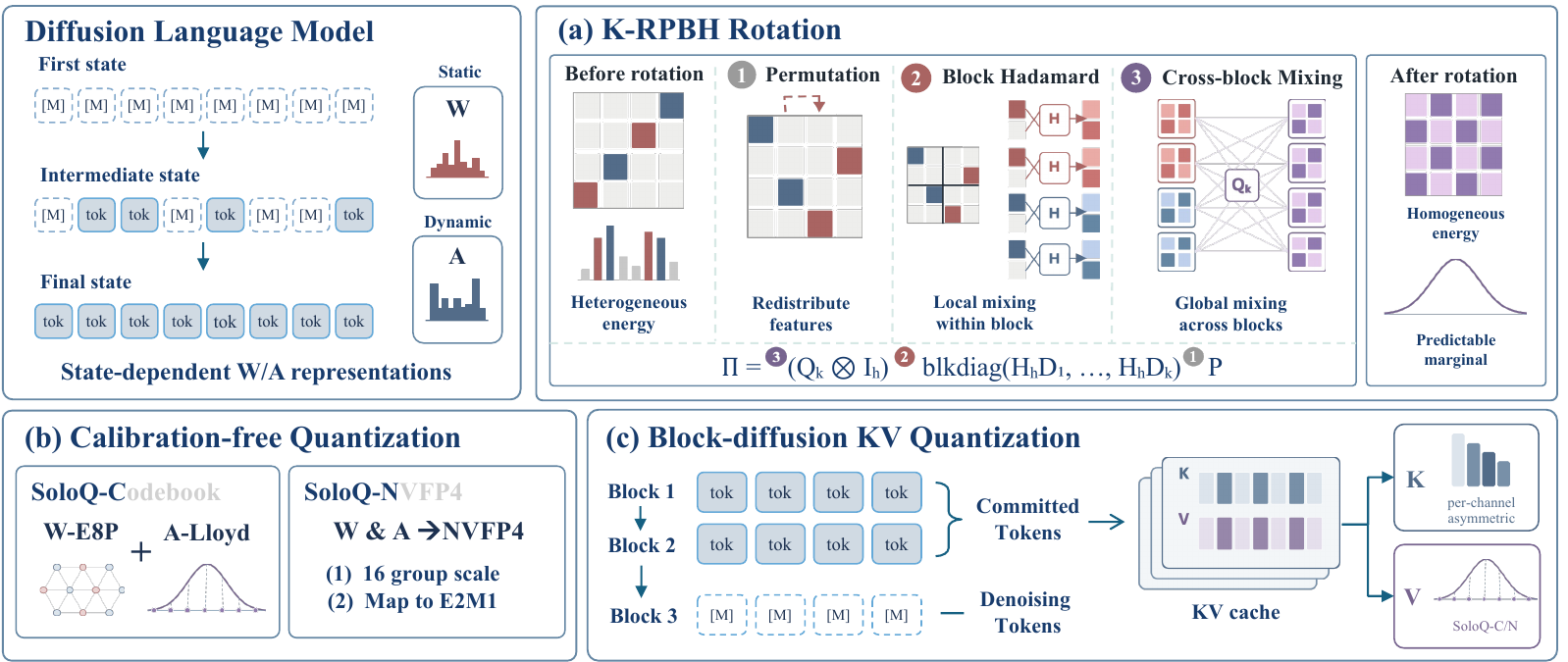}
\vspace{-6mm}
\caption{
\textbf{Overview of \soloq.}
(a) K-RPBH combines permutation, block-Hadamard, and cross-block mixing to transform heterogeneous representations into a predictable space.
(b) \soloq-C and \soloq-N enable calibration-free weight--activation quantization using distribution-matched codebooks and NVFP4, respectively.
(c) For block-diffusion models, \soloq applies commit-time KV quantization, keeping the denoising block in full precision.
}
\label{fig:overall}
\vspace{-3mm}
\end{figure}

\vspace{-2mm}
Rather than tracking activation distributions that shift across masking states and denoising steps, \soloq transforms representations into a predictable rotated space for low-bit quantization.
It applies K-RPBH for global mixing (Section~\ref{sec:krpbh}), followed by distribution-matched codebooks (\textsc{SoloQ-C}) or hardware-native NVFP4 (\textsc{SoloQ-N}) (Section~\ref{sec:calibfree_wa}) and a lightweight scale-only correction (Section~\ref{sec:unbiased_rescaling}).
For block-diffusion models, \soloq further extends this design to persistent KV states through commit-time quantization (Section~\ref{sec:kv_quant}).
Figure~\ref{fig:overall} illustrates the overall pipeline.

\vspace{-3mm}
\subsection{K-RPBH Rotation}
\label{sec:krpbh}
\vspace{-2mm}
Rotation plays a key role in calibration-free quantization by mapping normalized
representations to a predictable, approximately Gaussian coordinate distribution,
as shown in Eq.~\ref{eq:fixed_marginal}.
RPBH, introduced in OrbitQuant~\citep{lee2026orbitquant}, achieves this efficiently, but has two limitations for heavy-tailed dLLM representations.
\textbf{(1) Outlier-sensitive block balancing:}
RPBH relies on the initial permutation to balance the energy assigned to each block, and its variance-concentration guarantee weakens as the largest input coordinate grows.
\textbf{(2) No cross-block correction:}
Once the permutation is fixed, each Hadamard transform operates independently within its block, so any residual block-energy imbalance cannot be redistributed across blocks.
This effect is particularly relevant for dimensions composed of many Hadamard blocks, such as Dream's feed-forward network (FFN) dimension $d=18944=512\times37$.

To address these limitations, we introduce the Kronecker-extended randomized
permuted block-Hadamard (K-RPBH) rotation,
\begin{equation}
\mathbf{\Pi}
=
(\mathbf{Q}_k \otimes \mathbf{I}_h)
\operatorname{blkdiag}
\left(
\mathbf{H}_h\mathbf{D}_1,\ldots,
\mathbf{H}_h\mathbf{D}_k
\right)
\mathbf{P},
\label{eq:krpbh}
\end{equation}
where $d=kh$ with $h$ a power of two, $\mathbf{H}_h\in\mathbb{R}^{h\times h}$ is the normalized
Walsh--Hadamard matrix, $\mathbf{D}_i\in\mathbb{R}^{h\times h}$ is a diagonal
matrix with i.i.d.\ Rademacher entries, $\mathbf{P}\in\mathbb{R}^{d\times d}$
is a uniformly random permutation matrix, $\mathbf{Q}_k\in\mathbb{R}^{k\times k}$
is a Haar-random orthogonal cross-block mixer, and
$\mathbf{I}_h\in\mathbb{R}^{h\times h}$ is the identity matrix.
The $\mathbf{Q}_k\otimes\mathbf{I}_h$ stage mixes corresponding
coordinates across all $k$ blocks, directly redistributing the residual block
statistics left by RPBH.
When $k=1$, K-RPBH reduces to RPBH.

\begin{proposition}[Cross-block variance averaging]
\label{prop:krpbh_variance}
Let $x\in\mathbb{R}^{d}$ be a unit vector with $d=kh$, and let
$M_j=\|(\mathbf{P}x)^{(j)}\|_2^2$ denote the energy of the $j$-th permuted block, with $\sum_{j=1}^{k}M_j=1$.
For the K-RPBH output $\hat z$, let $\hat z_{\ell}^{(r)}$ denote the $\ell$-th coordinate of output block $r$.
Conditioned on $\mathbf{P}$ and $\mathbf{Q}_k$,
\begin{equation}
\operatorname{Var}_{\mathbf D}
\!\left(
\hat z_{\ell}^{(r)}
\mid \mathbf{P},\mathbf{Q}_k
\right)
=
\frac{1}{h}
\sum_{j=1}^{k}Q_{rj}^{2}M_j,
\label{eq:krpbh_var}
\end{equation}
where the variance is taken over the independent Rademacher sign matrices
$\mathbf{D}_1,\ldots,\mathbf{D}_k$.
If $\mathbf{Q}_k$ is Haar-random orthogonal, then
\begin{equation}
\mathbb{E}_{\mathbf Q_k}
\!\left[
\operatorname{Var}_{\mathbf D}
\!\left(
\hat z_{\ell}^{(r)}
\mid \mathbf{P},\mathbf{Q}_k
\right)
\right]
=
\frac{1}{d}.
\label{eq:krpbh_expected_var}
\end{equation}
\end{proposition}
The proof is provided in Appendix~\ref{app:proof_krpbh}.
Figure~\ref{fig:qualitative} empirically validates this effect on representative FFN
down-projection activations from Dream-7B~\citep{ye2025dream} and Nemotron-Labs-Diffusion~\citep{fu2026nemotron}. K-RPBH brings the rotated coordinate marginal substantially closer to the Gaussian target and reduces the block-to-block conditional-variance fluctuation left by RPBH. This improved distributional matching translates directly to lower quantization error while retaining the efficiency of a structured rotation.
K-RPBH remains substantially cheaper than the Kronecker transforms used in QuIP~\citep{chee2023quip}, and requires fewer transform operations than QuaRot~\citep{ashkboos2024quarot} on supported dimensions, as quantified in Figure~\ref{fig:qualitative}(c,f).

\begin{figure}[t]
\centering
\includegraphics[width=\linewidth]{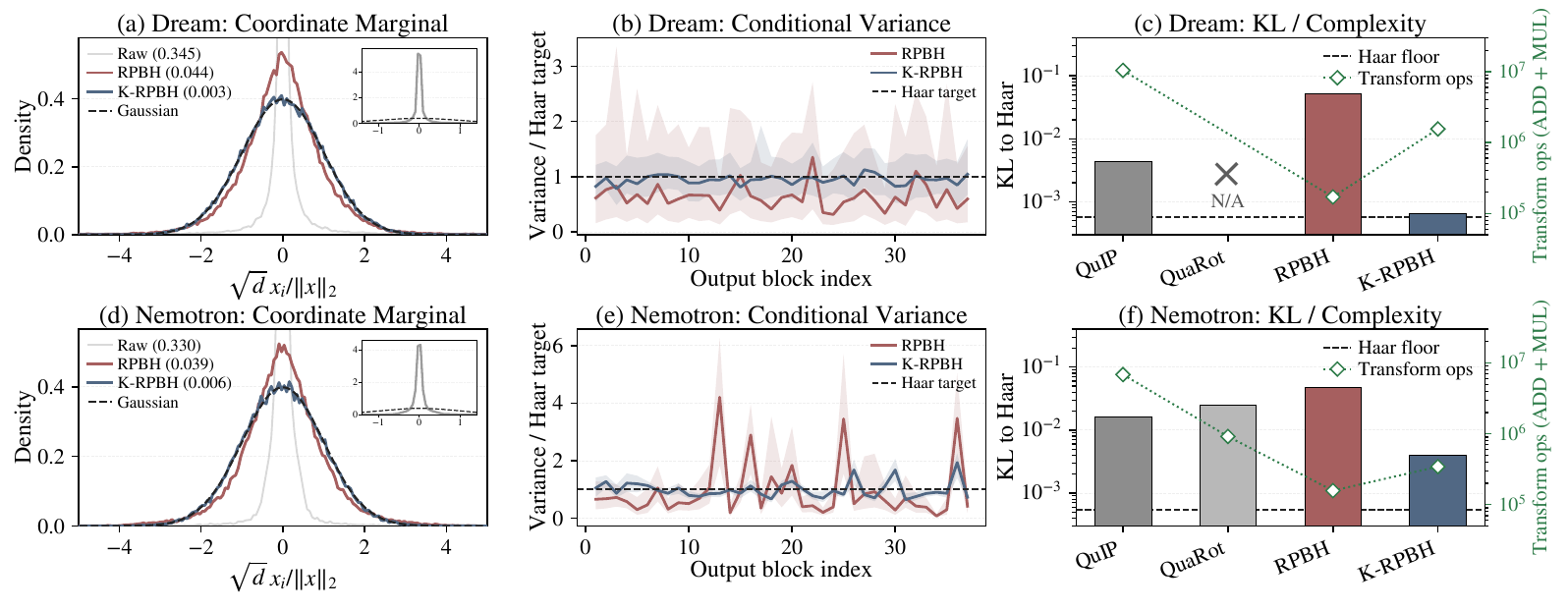}
\vspace{-8mm}
\caption{
\textbf{Effect of K-RPBH rotation on dLLM activations.}
Top row shows Dream-7B layer-1 FFN down projection and bottom row shows Nemotron layer-2 FFN down projection.
(a,d) Coordinate marginals before and after rotation.
(b,e) Conditional coordinate variance across output blocks, normalized by the Haar target $1/d$.
(c,f) KL divergence to the Haar reference versus transform complexity. N/A denotes an unsupported transform dimension for QuaRot~\citep{ashkboos2024quarot}.
}
\label{fig:qualitative}
\vspace{-4mm}
\end{figure}

\vspace{-2mm}

\subsection{Calibration-Free Weight--Activation Quantization} \label{sec:calibfree_wa} \vspace{-2mm} The predictable coordinate distribution produced by K-RPBH enables \soloq to quantize weights and activations without calibration data. Consider a linear layer $y=\mathbf{W}x$, where $\mathbf{W}\in\mathbb{R}^{m\times d}$ and $x\in\mathbb{R}^{d}$. Following prior rotation-based quantization~\citep{ashkboos2024quarot,lee2026orbitquant}, \soloq rotates both operands into the same K-RPBH basis and applies a quantizer $Q(\cdot)$: \begin{equation} \mathbf{W}'=\mathbf{W}\mathbf{\Pi}^{\top}, \qquad x'=\mathbf{\Pi}x, \qquad Q(\mathbf{W}')Q(x') \approx \mathbf{W}'x'
=\mathbf{W}x,
\label{eq}
\end{equation}
where $\mathbf{\Pi}$ is the K-RPBH transform in Equation~\ref{eq:krpbh}.
The weight rotation is performed offline, while the activation rotation is applied online.
\soloq then supports either distribution-matched codebook quantization or hardware-native NVFP4 on the shared rotated representation.


\paragraph{Distribution-matched codebooks (\soloq-C).}
For codebook quantization, \soloq-C first computes scales from the rotated operands and normalizes them before quantization.
Let $w_i'$ denote the $i$-th row of $\mathbf{W}'$.
We define
\begin{equation}
s_x=\|x'\|_2,
\qquad
s_{W,i}=\|w_i'\|_2,
\qquad
\tilde{x}=\frac{x'}{s_x},
\qquad
\tilde{w}_i=\frac{w_i'}{s_{W,i}}.
\label{eq:cb_normalization}
\end{equation}
The normalized coordinates are then quantized with distribution-matched codebooks,
\begin{equation}
\hat{x}'_{\mathrm{CB}}
=
s_x Q_{\mathrm{Lloyd}}(\tilde{x}),
\qquad
\hat{w}'_{i,\mathrm{CB}}
=
s_{W,i} Q_{\mathrm{E8P}}(\tilde{w}_i),
\label{eq:cb_quantization}
\end{equation}
where $Q_{\mathrm{Lloyd}}$ is a Lloyd--Max scalar quantizer for activations~\citep{zandieh2026turboquant}, and $Q_{\mathrm{E8P}}$ is an E8P-based lattice quantizer for weights~\citep{tseng2024quip}, with residual stages for higher bit-widths. We use different quantizers because weights are quantized offline, where E8P search is affordable, whereas activations require lightweight runtime quantization. Both are derived from the predictable post-rotation distribution without calibration data or activation statistics.

\vspace{-3mm}
\paragraph{Hardware-native NVFP4 (\soloq-N).}
\soloq also supports NVFP4 on the rotated representation, using the fixed E2M1
format with dynamic tensor- and block-wise scaling. For a rotated weight or
activation tensor $z$, we compute the global amplitude
$a_{\mathrm{g}}=\max_i |z_i|$ and block amplitude
$a_j=\max_{i\in\mathcal{G}_j}|z_i|$ over groups of 16 coordinates. The
corresponding scales are
\begin{equation}
s_{\mathrm{g}} = f_{\mathrm{g}}(a_{\mathrm{g}}),
\qquad
s_j = f_{\mathrm{E4M3}}(a_j),
\label{eq:nvfp4_scale}
\end{equation}
and the quantized reconstruction is
\begin{equation}
\hat{z}_{\mathrm{NVFP4}}^{(j)}
=
s_{\mathrm{g}}s_j\,
Q_{\mathrm{E2M1}}
\left(
\frac{z^{(j)}}{s_{\mathrm{g}}s_j}
\right),
\label{eq:nvfp4_quant}
\end{equation}
where $Q_{\mathrm{E2M1}}$ maps values to the representable E2M1 FP4 set.
All scales are derived directly from $z$, without calibration data,
pre-collected activation statistics, or learned parameters.

Thus, \soloq supports two calibration-free paths: \soloq-C with distribution-matched codebooks and \soloq-N with hardware-native NVFP4, both using the predictable rotated representation.

\vspace{-2mm}
\subsection{Scale-Only Bias Correction}
\label{sec:unbiased_rescaling}
\vspace{-2mm}
Low-bit quantization can distort both the direction and magnitude of a
representation, particularly with centroid-based quantizers such as
Lloyd--Max and, to a lesser extent, finite-precision formats such as NVFP4.
Related scalar corrections have been used for unbiased compression in
DRIVE~\citep{vargaftik2021drive} and EDEN~\citep{vargaftik2022eden}, and for
unbiased inner-product estimation in RaBitQ~\citep{gao2024rabitq}.
TurboQuant~\citep{zandieh2026turboquant} instead uses a residual QJL sketch,
requiring auxiliary information and a separate correction path.
\soloq uses a scale-only bias correction to restore the projection along the
original vector. Let $c_{\mathrm{bc}}(z)$ denote the corresponding
bias-correction scale for a vector $z$. We define
\begin{equation}
c_{\mathrm{bc}}(z)
=
\frac{\|z\|_2^2}{\langle q(z),z\rangle},
\qquad
\left\langle c_{\mathrm{bc}}(z)q(z),z\right\rangle
=
\|z\|_2^2.
\label{eq:bc_scale}
\end{equation}
Thus, the component along the original vector is exactly restored.
Equivalently, decompose
$q(z)=\alpha z+e_{\perp}$, where
$\alpha=\langle q(z),z\rangle/\|z\|_2^2$ and
$e_{\perp}\perp z$ denotes the orthogonal quantization error.
Then
$c_{\mathrm{bc}}(z)q(z)=z+e_{\perp}/\alpha$.
Unlike residual-based corrections, $c_{\mathrm{bc}}$ requires no auxiliary
payload or additional quantized bits. It is absorbed into the existing
per-token activation or per-row weight scale, leaving the underlying
low-bit linear operation unchanged.

\vspace{-2mm}
\subsection{Block-Diffusion KV-Cache Quantization}
\label{sec:kv_quant}
\vspace{-2mm}
Full-sequence dLLMs such as LLaDA~\citep{nie2026large,zhu2026llada} and Dream~\citep{ye2025dream}
recompute KV states throughout denoising and therefore do not maintain a persistent KV-cache.
In contrast, block-diffusion models~\citep{bie2026llada21speedingtextdiffusion,fu2026nemotron,wu2026fast}
reuse the KV-cache of previously committed blocks, causing the persistent cache to grow with context length.
As shown in Figure~\ref{fig:motivation}(a), this growth creates a significant memory bottleneck at long context lengths.

\soloq quantizes the KV-cache only when a block is committed, as illustrated in
Figure~\ref{fig:overall}(c).
The active denoising block remains in full precision, avoiding repeated
quantization across denoising steps.
Keys use KIVI-style per-channel asymmetric quantization~\citep{liu2024kivi},
while values follow the \soloq activation path with K-RPBH and either
Lloyd--Max or NVFP4 quantization.
This commit-time design avoids repeatedly quantizing the KV states of the actively denoised block while compressing only persistent states.
The overall algorithm of \soloq is shown in Appendix~\ref{Apix:algo}.

\begin{table*}[t]
\caption{
W4A4 results on full-sequence diffusion language models.
Values are task accuracy (\%) and their average.
\soloq-C and \soloq-N denote the codebook and NVFP4 variants, respectively.
\textbf{Bold} and \underline{underlined} indicate the best and second-best quantized results within each model.
}
\vspace{-2mm}
\label{tab:full_dllm}
\small
\centering
\resizebox{\textwidth}{!}{%
\begin{tabular}{clccccccccccc}
\toprule
Model & Method & Calibration-Free
& Truth.
& ARC-C
& Hella.
& Wino.
& PIQA
& MMLU
& C-EVAL
& Human.
& GSM8K
& Avg. \\
\midrule

\multirow{9}{*}{LLaDA-Base-8B}
& FP16 & -
& 47.45 & 44.03 & 54.09 & 74.82 & 74.81 & 64.15 & 69.99 & 31.71 & 69.52 & 58.95 \\
\cdashline{2-13}
& RTN & \cmark
& 40.45 & 41.83 & 45.40 & 64.72 & 67.95 & 49.26 & 57.95 & 14.02 & 16.56 & 44.23 \\
& AWQ & \xmark
& 40.87 & 42.92 & 46.14 & 66.88 & 69.43 & 51.22 & 58.43 & 20.10 & 36.88 & 48.09 \\
& QuaRot+GPTQ &\xmark
& 42.53 & 44.20 & 49.76 & 69.85 & 70.75 & 55.96 & 56.32 & 25.33 & 44.57 & 51.03 \\
& DLLMQuant+ &\xmark
& 41.53 & 43.44 & 46.51 & 67.87 & 70.12 & 51.72 & 59.38 & 22.13 & 40.66 & 49.26 \\
& DLLMQuant++ & \xmark
& 43.53 & 44.18 & 51.00 & 71.85 & 73.94 & 57.77 & 61.22 & 28.92 & 56.25 & 54.29 \\
& STaR-Quant &\xmark
& \textbf{49.12}
& \underline{44.23}
& 52.75
& 72.92
& 73.85
& 62.95
& 64.56
& \textbf{35.98}
& 57.29
& 57.07 \\
\cdashline{2-13}
& \cellcolor{gray!25}\soloq-C 
& \cellcolor{gray!25}\cmark
& \cellcolor{gray!25}45.22
& \cellcolor{gray!25}43.52
& \cellcolor{gray!25}\underline{52.94}
& \cellcolor{gray!25}\underline{74.51}
& \cellcolor{gray!25}\underline{74.16}
& \cellcolor{gray!25}\underline{64.11}
& \cellcolor{gray!25}\underline{67.53}
& \cellcolor{gray!25}\underline{29.88}
& \cellcolor{gray!25}\textbf{67.78}
& \cellcolor{gray!25}\textbf{57.74} \\
& \cellcolor{gray!25}\soloq-N
& \cellcolor{gray!25}\cmark
& \cellcolor{gray!25}\underline{45.33}
& \cellcolor{gray!25}\textbf{45.22}
& \cellcolor{gray!25}\textbf{53.13}
& \cellcolor{gray!25}\textbf{74.59}
& \cellcolor{gray!25}\textbf{74.48}
& \cellcolor{gray!25}\textbf{64.55}
& \cellcolor{gray!25}\textbf{67.98}
& \cellcolor{gray!25}25.61
& \cellcolor{gray!25}\underline{67.48}
& \cellcolor{gray!25}\underline{57.60} \\

\midrule

\multirow{9}{*}{LLaDA-1.5-8B}
& FP16 & -
& 47.20 & 87.46 & 75.02 & 72.93 & 74.86 & 65.47 & 66.05 & 50.61 & 83.47 & 69.23 \\
\cdashline{2-13}
& RTN & \cmark
& 39.51 & 81.77 & 56.82 & 65.27 & 66.91 & 48.88 & 58.96 & 23.44 & 36.56 & 53.12 \\
& AWQ & \xmark
& 40.96 & 82.22 & 66.84 & 67.93 & 69.32 & 51.12 & 60.03 & 30.07 & 57.95 & 58.49 \\
& QuaRot+GPTQ & \xmark
& 43.21 & 84.23 & 65.34 & 69.55 & 70.17 & 56.23 & 57.66 & 37.33 & 65.86 & 61.06 \\
& DLLMQuant+ & \xmark
& 42.14 & 83.38 & 70.09 & 68.69 & 70.22 & 51.63 & 61.28 & 32.44 & 59.55 & 59.94 \\
& DLLMQuant++ & \xmark
& 43.87 & 84.18 & 69.20 & \underline{71.74} & 73.58 & 57.27 & 60.04 & 44.58 & 74.33 & 64.31 \\
& STaR-Quant & \xmark
& \textbf{48.77}
& 85.12
& 72.43
& \textbf{72.09}
& 73.36
& 63.29
& 64.34
& 46.53
& 76.45
& 66.93 \\
\cdashline{2-13}
& \cellcolor{gray!25}\soloq-C
& \cellcolor{gray!25}\cmark
& \cellcolor{gray!25}\underline{48.10}
& \cellcolor{gray!25}\textbf{86.44}
& \cellcolor{gray!25}\underline{73.01}
& \cellcolor{gray!25}71.59
& \cellcolor{gray!25}\underline{74.65}
& \cellcolor{gray!25}\underline{64.60}
& \cellcolor{gray!25}\underline{64.71}
& \cellcolor{gray!25}\underline{46.95}
& \cellcolor{gray!25}\underline{80.29}
& \cellcolor{gray!25}\underline{67.82} \\
& \cellcolor{gray!25}\soloq-N
& \cellcolor{gray!25}\cmark
& \cellcolor{gray!25}48.09
& \cellcolor{gray!25}\underline{86.10}
& \cellcolor{gray!25}\textbf{73.86}
& \cellcolor{gray!25}71.67
& \cellcolor{gray!25}\textbf{74.76}
& \cellcolor{gray!25}\textbf{64.70}
& \cellcolor{gray!25}\textbf{65.38}
& \cellcolor{gray!25}\textbf{50.00}
& \cellcolor{gray!25}\textbf{81.80}
& \cellcolor{gray!25}\textbf{68.48} \\

\midrule

\multirow{9}{*}{Dream-7B$\dagger$}
& FP16 & -
& 43.96 & 59.98 & 73.73 & 74.43 & 75.46 & 71.85 & 62.48 & 59.15 & 77.79 & 66.54 \\
\cdashline{2-13}
& RTN & \cmark
& 41.25 & 54.34 & 58.32 & 64.52 & 66.26 & 51.51 & 49.89 & 28.90 & 30.82 & 49.53 \\
& AWQ & \xmark
& 43.66 & 57.82 & 65.57 & 67.13 & 68.96 & 55.50 & 53.27 & 33.14 & 48.98 & 54.89 \\
& QuaRot+GPTQ & \xmark
& 47.58 & 58.18 & 67.13 & 70.05 & 70.36 & \textbf{69.50} & 53.19 & 34.48 & 59.20 & 58.85 \\
& DLLMQuant+ & \xmark
& 44.14 & 58.38 & 66.93 & 68.66 & 69.53 & 57.63 & 54.21 & 35.12 & 51.14 & 56.19 \\
& DLLMQuant++ & \xmark
& \underline{47.86}
& \underline{59.43}
& 70.14
& 71.54
& 72.09
& \textbf{69.50}
& 55.89
& 44.50
& 66.17
& 61.90 \\
& STaR-Quant & \xmark
& \textbf{48.72}
& \textbf{59.76}
& 71.32
& \textbf{72.98}
& 74.34
& \underline{69.37}
& \textbf{58.79}
& \underline{47.43}
& 69.57
& \textbf{63.59} \\
\cdashline{2-13}
& \cellcolor{gray!25}\soloq-C
& \cellcolor{gray!25}\cmark
& \cellcolor{gray!25}41.99
& \cellcolor{gray!25}55.89
& \cellcolor{gray!25}\underline{71.62}
& \cellcolor{gray!25}71.35
& \cellcolor{gray!25}\textbf{74.76}
& \cellcolor{gray!25}68.86
& \cellcolor{gray!25}\underline{56.09}
& \cellcolor{gray!25}45.12
& \cellcolor{gray!25}\textbf{72.86}
& \cellcolor{gray!25}62.06 \\
& \cellcolor{gray!25}\soloq-N
& \cellcolor{gray!25}\cmark
& \cellcolor{gray!25}42.95
& \cellcolor{gray!25}56.06
& \cellcolor{gray!25}\textbf{72.05}
& \cellcolor{gray!25}\underline{72.22}
& \cellcolor{gray!25}\underline{74.54}
& \cellcolor{gray!25}68.97
& \cellcolor{gray!25}54.98
& \cellcolor{gray!25}\textbf{50.61}
& \cellcolor{gray!25}\underline{70.36}
& \cellcolor{gray!25}\underline{62.53} \\

\bottomrule
\end{tabular}%
}
\vspace{0.4mm}
\noindent
{\scriptsize
\raggedleft
\textsuperscript{$\dagger$} TruthfulQA results for DLLMQuant/STaR-Quant are not reproducible (not open-sourced). Their FP is 49.76 vs. 43.96 in our setup.
\par
}
\vspace{-2mm}
\end{table*}

\begin{table*}[t]
\caption{
W4A4 and W4A4KV4 results on block-diffusion language models.
Values are task accuracy (\%) and average.
\soloq-C and \soloq-N denote the codebook and NVFP4 variants.
\textbf{Bold} and \underline{underlined} mark the best and second-best quantized results per model and precision.
Cb-free denotes whether the method is calibration-free, and N/A denotes an inapplicable method.}
\vspace{-2mm}
\label{tab:block_dllm}
\small
\centering
\resizebox{\textwidth}{!}{%
\begin{tabular}{cclccccccccccc}
\toprule
Model & Precision & Method & Cb-free
& Human-B
& Human-P
& MBPP-B
& MBPP-P
& GSM8K
& MATH
& IFEval
& MMLU
& GPQA
& Avg. \\
\midrule

\multirow{9}{*}{Fast-dLLM v2-7B}
& FP16 & -- & --
& 65.85 & 59.76 & 61.90 & 52.65 & 84.15 & 60.30 & 61.18 & 66.54 & 30.80 & 60.35 \\
\cdashline{2-14}

& \multirow{5}{*}{W4A4}
& RTN
& \cmark
& 0.00 & 0.00 & 0.00 & 0.00 & 0.23 & 0.00 & 7.58 & 24.73 & 27.46 & 6.67 \\
&
& AWQ
& \xmark
& 0.00 & 0.00 & 0.00 & 0.00 & 0.00 & 0.00 & 9.06 & 24.98 & 25.45 & 6.61 \\
&
& QuaRot+GPTQ
& \xmark
& N/A & N/A & N/A & N/A & N/A & N/A & N/A & N/A & N/A & N/A \\
&
& \cellcolor{gray!25}\soloq-C
& \cellcolor{gray!25}\cmark
& \cellcolor{gray!25}\underline{58.54}
& \cellcolor{gray!25}\underline{53.05}
& \cellcolor{gray!25}\textbf{60.05}
& \cellcolor{gray!25}\textbf{50.53}
& \cellcolor{gray!25}\underline{80.74}
& \cellcolor{gray!25}\textbf{54.86}
& \cellcolor{gray!25}\textbf{59.89}
& \cellcolor{gray!25}\textbf{64.73}
& \cellcolor{gray!25}\underline{30.58}
& \cellcolor{gray!25}\textbf{57.00} \\
&
& \cellcolor{gray!25}\soloq-N
& \cellcolor{gray!25}\cmark
& \cellcolor{gray!25}\textbf{59.10}
& \cellcolor{gray!25}\textbf{56.10}
& \cellcolor{gray!25}\underline{57.40}
& \cellcolor{gray!25}\underline{48.90}
& \cellcolor{gray!25}\textbf{82.64}
& \cellcolor{gray!25}\underline{54.42}
& \cellcolor{gray!25}\underline{55.27}
& \cellcolor{gray!25}\underline{63.43}
& \cellcolor{gray!25}\textbf{33.71}
& \cellcolor{gray!25}\underline{56.77} \\

\cdashline{2-14}

& \multirow{2}{*}{W4A4KV4}
& \cellcolor{gray!25}\soloq-C
& \cellcolor{gray!25}\cmark
& \cellcolor{gray!25}\underline{58.54}
& \cellcolor{gray!25}\underline{52.44}
& \cellcolor{gray!25}\underline{57.67}
& \cellcolor{gray!25}\textbf{51.32}
& \cellcolor{gray!25}\textbf{79.98}
& \cellcolor{gray!25}\textbf{54.06}
& \cellcolor{gray!25}\textbf{56.38}
& \cellcolor{gray!25}\textbf{64.73}
& \cellcolor{gray!25}\underline{30.58}
& \cellcolor{gray!25}\underline{56.19} \\
&
& \cellcolor{gray!25}\soloq-N
& \cellcolor{gray!25}\cmark
& \cellcolor{gray!25}\textbf{59.15}
& \cellcolor{gray!25}\textbf{55.49}
& \cellcolor{gray!25}\textbf{57.94}
& \cellcolor{gray!25}\underline{49.21}
& \cellcolor{gray!25}\underline{79.91}
& \cellcolor{gray!25}\underline{53.78}
& \cellcolor{gray!25}\underline{55.08}
& \cellcolor{gray!25}\underline{63.43}
& \cellcolor{gray!25}\textbf{31.92}
& \cellcolor{gray!25}\textbf{56.21} \\

\midrule

\multirow{9}{*}{\shortstack{Nemotron-Labs\\Diffusion-8B}}
& FP16 & -- & --
& 84.76 & 78.66 & 82.28 & 69.31 & 92.95 & 85.10 & 97.27 & 78.17 & 39.51 & 78.67 \\
\cdashline{2-14}

& \multirow{5}{*}{W4A4}
& RTN
& \cmark
& 0.00 & 0.00 & 0.00 & 0.00 & 0.00 & 0.00 & 11.70 & 0.00 & 0.00 & 1.30 \\
&
& AWQ
& \xmark
& 0.00 & 0.00 & 0.00 & 0.00 & 0.00 & 0.00 & 11.53 & 0.00 & 0.00 & 1.28 \\
&
& QuaRot+GPTQ
& \xmark
& 68.30 & 64.60 & 69.58 & 58.99 & 79.76 & 65.70 & 53.15 & 59.27 & 19.20 & 59.39 \\
&
& \cellcolor{gray!25}\soloq-C
& \cellcolor{gray!25}\cmark
& \cellcolor{gray!25}\underline{75.61}
& \cellcolor{gray!25}\underline{70.73}
& \cellcolor{gray!25}\underline{75.66}
& \cellcolor{gray!25}\underline{63.49}
& \cellcolor{gray!25}\underline{89.92}
& \cellcolor{gray!25}\underline{81.10}
& \cellcolor{gray!25}\textbf{62.35}
& \cellcolor{gray!25}\underline{72.84}
& \cellcolor{gray!25}\textbf{35.94}
& \cellcolor{gray!25}\underline{69.74} \\
&
& \cellcolor{gray!25}\soloq-N
& \cellcolor{gray!25}\cmark
& \cellcolor{gray!25}\textbf{76.22}
& \cellcolor{gray!25}\textbf{72.56}
& \cellcolor{gray!25}\textbf{78.84}
& \cellcolor{gray!25}\textbf{67.46}
& \cellcolor{gray!25}\textbf{91.21}
& \cellcolor{gray!25}\textbf{81.12}
& \cellcolor{gray!25}\underline{60.85}
& \cellcolor{gray!25}\textbf{74.17}
& \cellcolor{gray!25}\underline{34.38}
& \cellcolor{gray!25}\textbf{70.76} \\

\cdashline{2-14}

& \multirow{2}{*}{W4A4KV4}
& \cellcolor{gray!25}\soloq-C
& \cellcolor{gray!25}\cmark
& \cellcolor{gray!25}\underline{73.17}
& \cellcolor{gray!25}\underline{69.51}
& \cellcolor{gray!25}\underline{77.78}
& \cellcolor{gray!25}\underline{65.61}
& \cellcolor{gray!25}\underline{91.28}
& \cellcolor{gray!25}\underline{80.62}
& \cellcolor{gray!25}\textbf{65.48}
& \cellcolor{gray!25}\textbf{73.98}
& \cellcolor{gray!25}\textbf{34.15}
& \cellcolor{gray!25}\underline{70.18} \\
&
& \cellcolor{gray!25}\soloq-N
& \cellcolor{gray!25}\cmark
& \cellcolor{gray!25}\textbf{75.00}
& \cellcolor{gray!25}\textbf{72.56}
& \cellcolor{gray!25}\textbf{80.16}
& \cellcolor{gray!25}\textbf{67.99}
& \cellcolor{gray!25}\textbf{91.66}
& \cellcolor{gray!25}\textbf{81.10}
& \cellcolor{gray!25}\underline{64.11}
& \cellcolor{gray!25}\underline{73.93}
& \cellcolor{gray!25}\underline{33.04}
& \cellcolor{gray!25}\textbf{71.06} \\

\bottomrule
\end{tabular}%
}
\vspace{-5.5mm}
\end{table*}

\vspace{-1.5mm}
\section{Experiments}
\vspace{-3mm}
\subsection{Experimental Setup}
\label{sec:exp_setup}
\vspace{-2mm}
\paragraph{Models and evaluation.}
We evaluate \soloq on both full-sequence and block-diffusion language models.
For full-sequence dLLMs, we use LLaDA-8B-Base~\citep{nie2026large},
LLaDA-1.5~\citep{zhu2026llada}, and Dream-7B~\citep{ye2025dream}.
For block-diffusion models, we evaluate Fast-dLLM v2~\citep{wu2026fast}
and Nemotron-Labs-Diffusion~\citep{fu2026nemotron}.
All accuracy evaluations are conducted on NVIDIA H200 GPUs.
We preserve each model's official architecture-specific evaluation stack rather
than enforcing a single evaluator across all models.
Detailed benchmark configurations are provided in
Appendix~\ref{app:eval_setup}.

\vspace{-2mm}
\paragraph{Baselines.}
For full-sequence dLLMs, we compare against standard LLM PTQ baselines including RTN, AWQ~\citep{lin2024awq}, QuaRot~\citep{ashkboos2024quarot}, and GPTQ~\citep{frantar2022gptq}, as well as dLLM-specific methods, DLLMQuant~\citep{xu2025dllmquant} and STaR-Quant~\citep{yan2026star}.
For block-diffusion models, where prior quantization results are limited, we compare against RTN, AWQ, and QuaRot under the same evaluation protocol. The additional comparison with Fair-Calib~\citep{huang2026fair} is presented in Appendix~\ref{fair-calib}.

\vspace{-3mm}
\subsection{Full-Sequence Diffusion Language Models}
\label{sec:full_sequence_results}
\vspace{-2mm}
Table~\ref{tab:full_dllm} reports W4A4 results on LLaDA-Base~\citep{nie2026large}, LLaDA-1.5~\citep{zhu2026llada}, and Dream-7B~\citep{ye2025dream}.
\soloq outperforms existing quantized baselines without calibration data.
\soloq-C achieves 57.74 average accuracy on LLaDA-Base, improving over DLLMQuant++ by 3.45 points, while \soloq-N reaches 68.48 on LLaDA-1.5, a 4.17-point gain.
On Dream-7B, \soloq-N achieves 62.53, versus 61.90 for DLLMQuant++.
Overall, both variants provide robust W4A4 quantization across generation, reasoning, and knowledge tasks on full-sequence dLLMs.

\vspace{-2mm}
\subsection{Block-Diffusion Language Models}
\label{sec:block_diffusion_results}
\vspace{-2mm}
To assess quantization in block-diffusion settings, we evaluate W4A4 and W4A4KV4 on Fast-dLLM v2~\citep{wu2026fast} and Nemotron-Labs-Diffusion~\citep{fu2026nemotron}, with results summarized in Table~\ref{tab:block_dllm}.
Unlike full-sequence dLLMs, block-diffusion models maintain persistent KV states, making KV-cache quantization an additional efficiency opportunity.
Most block-diffusion benchmarks are generation-based rather than likelihood-based, which can make them more sensitive to quantization error. As a result, RTN and AWQ collapse under W4A4 on both models, while QuaRot is not applicable to Fast-dLLM v2 because its rotation cannot support the model dimensions.
Across both models, \soloq maintains strong accuracy under W4A4 and W4A4KV4, demonstrating effective calibration-free quantization of weights, activations, and persistent KV states.
Appendix~\ref{sink} analyzes attention-sink~\citep{xiao2024efficient, rulli2026attention}
behavior under KV-cache quantization.

\begin{figure}[t]
\centering
\includegraphics[width=\linewidth]{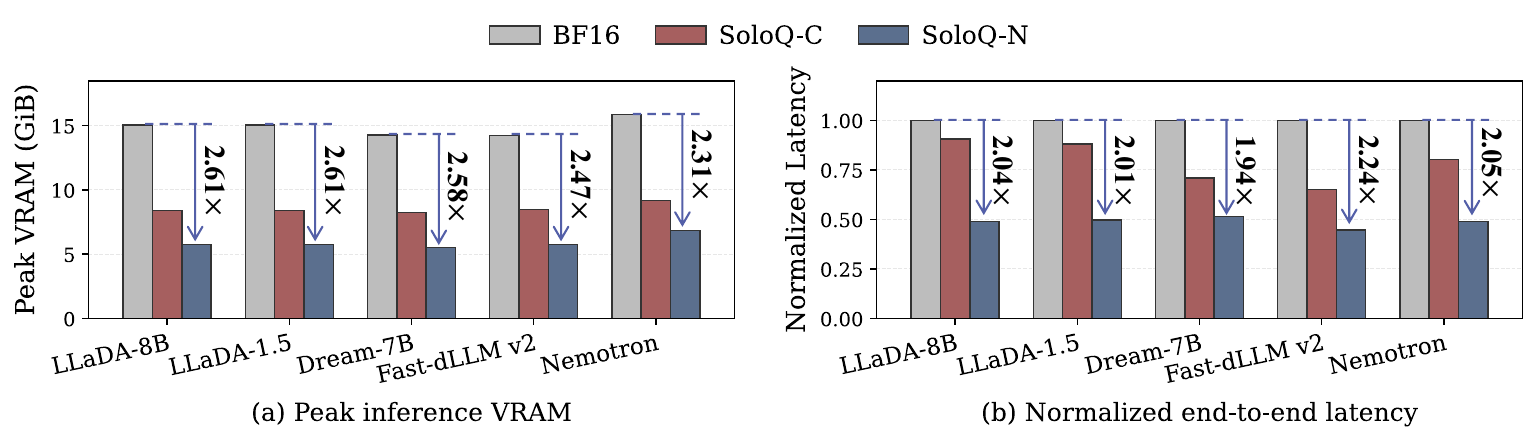}
\vspace{-7mm}
\caption{
\textbf{End-to-end efficiency of \soloq.}
(a) Peak inference VRAM and (b) normalized generation latency across five dLLMs.
\soloq-C and \soloq-N are evaluated on H200 and RTX Pro 6000 GPUs, respectively, against their corresponding BF16 baselines.
}

\label{fig:memory}
\vspace{-2mm}
\end{figure}

\begin{table*}[t]
\centering

\begin{minipage}[t]{0.48\textwidth}
\vspace{0pt}
\centering

\captionof{table}{Rotation ablation under \soloq-N, comparing task accuracy and rotation latency. Lat. denotes latency. N/A indicates dimensions unsupported by QuaRot.}
\label{tab:rotation_ablation}
\vspace{-3mm}
\scriptsize
\setlength{\tabcolsep}{2.6pt}
\renewcommand{\arraystretch}{1.05}

\resizebox{\linewidth}{!}{%
\begin{tabular}{lcccc}
\toprule
& \multicolumn{2}{c}{LLaDA-1.5}
& \multicolumn{2}{c}{Fast-dLLM v2} \\
\cmidrule(lr){2-3}
\cmidrule(lr){4-5}
Rotation
& Hella./MMLU
& Lat.
& Human+/GPQA
& Lat. \\
& (\%)
& (ms)
& (\%)
& (ms) \\
\midrule

QuIP
& 73.29/63.17
& 0.086
& 55.40/30.58
& 10.495 \\

QuaRot
& 73.41/63.19
& 0.025
& N/A
& N/A \\

Haar
& 73.79/63.51
& 0.559
& \textbf{57.90/33.48}
& 98.135 \\

RPBH
& 73.08/63.48
& 0.031
& 53.00/32.81
& 2.447 \\

\cellcolor{gray!25}K-RPBH
& \cellcolor{gray!25}\textbf{73.86/64.70}
& \cellcolor{gray!25}0.048
& \cellcolor{gray!25}\underline{56.10/33.71}
& \cellcolor{gray!25}5.779 \\

\bottomrule
\end{tabular}%
}
\end{minipage}
\hfill
\begin{minipage}[t]{0.48\textwidth}
\vspace{0pt}
\centering

\captionof{table}{
Component ablation of K-RPBH on Nemotron.
$P$ denotes randomized permutation and $Q_k$ cross-block mixing.
Lower KL and energy imbalance indicate better distribution matching and global mixing.
}
\label{tab:krpbh_component}
\vspace{1mm}
\scriptsize
\setlength{\tabcolsep}{3.0pt}
\renewcommand{\arraystretch}{1.35}

\resizebox{\linewidth}{!}{%
\begin{tabular}{lcccc}
\toprule
Variant
& $P$
& $Q_k$
& KL $\downarrow$
& Energy Imbalance $\downarrow$ \\
\midrule

Block-RHT
& \xmark
& \xmark
& 0.11396
& 67.89\% \\

Block-RHT + $P$
& \cmark
& \xmark
& 0.04683
& 45.64\% \\

Block-RHT + $Q_k$
& \xmark
& \cmark
& 0.00682
& 36.24\% \\

\cellcolor{gray!25}K-RPBH (ours)
& \cellcolor{gray!25}\cmark
& \cellcolor{gray!25}\cmark
& \cellcolor{gray!25}\textbf{0.00413}
& \cellcolor{gray!25}\textbf{22.12\%} \\

\bottomrule
\end{tabular}
}
\end{minipage}
\vspace{-5mm}
\end{table*}

\vspace{-2mm}
\subsection{Memory and Latency Efficiency}
\label{sec:efficiency}
\vspace{-2mm}
We measure the end-to-end inference efficiency of \soloq using optimized GPU kernel implementations.
\soloq-C uses 8-bit Tensor Core kernels on NVIDIA H200 GPUs, while \soloq-N uses native NVFP4 kernels on an NVIDIA RTX Pro 6000.
As shown in Figure~\ref{fig:memory}, \soloq-N achieves a $2.31\times$--$2.61\times$ reduction in peak memory across five dLLMs, primarily through 4-bit weight storage and, for block-diffusion models, KV-cache compression.
It also achieves $1.94\times$--$2.24\times$ end-to-end speedup over BF16 across the evaluated workloads.
For Nemotron, whose EOS trajectory changes after quantization, we compare the same BF16-recorded sequence of 75 forward passes and obtain a $2.05\times$ speedup.
We further implement \soloq on an FPGA, as detailed in Appendix~\ref{fpga}.

\vspace{-4mm}
\section{Analysis}
\vspace{-3mm}
\subsection{Rotation Structure Comparison}
\vspace{-3mm}
We first compare alternative rotations, QuIP~\citep{chee2023quip}, QuaRot~\citep{ashkboos2024quarot}, Haar~\citep{zandieh2026turboquant}, and RPBH~\citep{lee2026orbitquant}, under the same \soloq-N NVFP4 setting, as shown in Table~\ref{tab:rotation_ablation}. Rotation latency is measured for a single activation-rotation invocation on an NVIDIA H200 under each model's native evaluation workload. Dense Haar provides strong accuracy but incurs substantially higher cost. K-RPBH retains comparable accuracy while reducing rotation latency from 98.135\,ms to 5.779\,ms on Fast-dLLM v2, and from 0.559\,ms to 0.048\,ms on LLaDA-1.5. Compared with RPBH, K-RPBH consistently improves downstream accuracy, showing that cross-block mixing provides a better accuracy--efficiency trade-off.

\vspace{-2mm}
\subsection{K-RPBH Component Ablation}

\vspace{-2mm}

We next isolate the contributions of the randomized permutation $P$ and cross-block mixer $Q_k$ in K-RPBH (Eq.~\ref{eq:krpbh}), as shown in Table~\ref{tab:krpbh_component}.
For a token representation $x\in\mathbb{R}^d$ partitioned into $k$ blocks of size $h=d/k$, we define the energy of block $j$ after permutation as
$E_j(x)=\|(Px)^{(j)}\|_2^2$, whose uniform target is $\|x\|_2^2/k$.
We measure block-energy imbalance as the median absolute relative deviation from this target,
$\operatorname{Median}_{x,j}\!\left|kE_j(x)/\|x\|_2^2-1\right|$.
For K-RPBH, the corresponding block energies are computed after cross-block mixing by $Q_k$.
Lower imbalance indicates more uniform energy across blocks, while lower KL divergence indicates a closer match to the target post-rotation marginal.
Starting from Block-RHT, adding $P$ reduces KL from 0.114 to 0.0468 and energy imbalance from 67.89\% to 45.64\%.
Adding only $Q_k$ is more effective at redistribution, reducing KL to 0.0068 and imbalance to 36.24\%.
Combining both further improves the metrics to 0.0041 and 22.12\%, respectively, confirming their complementary benefits.

\begin{table*}[t]
\centering

\begin{minipage}[t]{0.58\textwidth}
\vspace{0pt}
\centering

\includegraphics[
    width=\linewidth
]{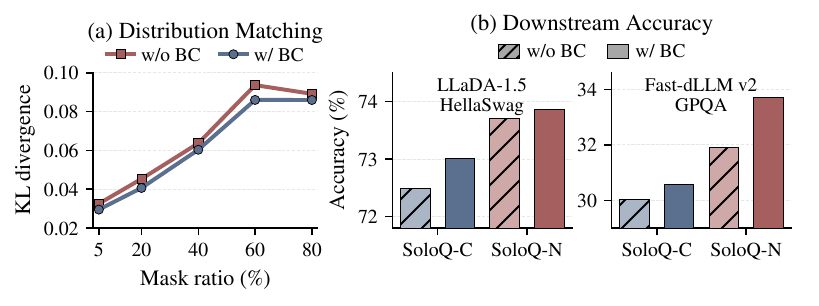}

\vspace{-2.5mm}
\captionof{figure}{
Effect of scale-only bias correction (BC).
(a) KL divergence between full-precision and quantized logits across token mask ratios on LLaDA-1.5.
(b) Downstream accuracy with and without BC.
}

\label{fig:ub_ablation}

\end{minipage}
\hfill
\begin{minipage}[t]{0.40\textwidth}
\vspace{0pt}
\centering

\captionof{table}{
Comparison of 2-bit KV-cache quantization methods on Fast-dLLM v2 with weights and activations kept in FP16.
Only the KV-cache is quantized, and BPE denotes its bit-per-entry.
}
\label{tab:kv_cache_comparison}
\vspace{-1.5mm}
\scriptsize
\setlength{\tabcolsep}{3.0pt}
\renewcommand{\arraystretch}{1.20}

\resizebox{\linewidth}{!}{%
\begin{tabular}{lcccc}
\toprule
Method
& BPE
& GSM8K (\%)
& IFEval (\%)
& Avg. (\%) \\
\midrule

FP16
& 16.00
& 84.15
& 61.18
& 72.67 \\
\cdashline{1-5}
KIVI
& 3.00
& 59.29
& 41.22
& 50.26 \\

OScar
& 3.03
& 52.39
& 40.85
& 46.62 \\

TurboQuant
& 3.18
& 0.30
& 7.76
& 4.03 \\

\cellcolor{gray!25}\soloq-C
& \cellcolor{gray!25}3.00
& \cellcolor{gray!25}\textbf{60.96}
& \cellcolor{gray!25}\textbf{45.66}
& \cellcolor{gray!25}\textbf{53.31} \\

\cellcolor{gray!25}\soloq-N
& \cellcolor{gray!25}3.00
& \cellcolor{gray!25}\textbf{60.65}
& \cellcolor{gray!25}\textbf{46.40}
& \cellcolor{gray!25}\textbf{53.53} \\

\bottomrule
\end{tabular}%
}

\end{minipage}
\vspace{-5mm}
\end{table*}
\vspace{-2mm}

\subsection{Effect of Scale-Only Bias Correction}
\vspace{-2mm}

We further study the effect of the scale-only bias correction in Figure~\ref{fig:ub_ablation}. Across all input-token masking ratios on LLaDA-1.5, the correction reduces the KL divergence between the full-precision and quantized logit distributions, indicating improved distribution matching. This translates to higher downstream accuracy for both \soloq-C and \soloq-N: HellaSwag improves from 72.49/73.71 to 73.01/73.86 on LLaDA-1.5, while GPQA improves from 30.03/31.92 to 30.58/33.71 on Fast-dLLM v2. Overall, the correction improves quantization fidelity with negligible overhead.

\vspace{-2mm}

\subsection{KV-Cache Quantization Comparison}
\vspace{-2mm}

Finally, we compare \soloq with existing KV-cache quantization methods using
2-bit KV quantization at approximately 3 bits-per-entry (BPE), including
scale and metadata overhead, in Table~\ref{tab:kv_cache_comparison}, while
keeping weights and activations in FP16.
\soloq-C and \soloq-N achieve average accuracies of 53.31 and 53.53,
outperforming KIVI~\citep{liu2024kivi}, OScar~\citep{su2026oscar}, and
TurboQuant~\citep{zandieh2026turboquant}.
\soloq-C performs best on GSM8K, while \soloq-N achieves the highest IFEval
accuracy, highlighting their complementary strengths.
These results show that \soloq effectively compresses the KV cache while
preserving downstream accuracy.

\vspace{-3mm}
\section{Conclusion}
\vspace{-3mm}
We present \soloq, a calibration-free quantization framework for diffusion language models. By transforming weights and activations into a predictable rotated representation, \soloq enables data-independent low-bit quantization without calibration samples or pre-collected activation statistics. The framework supports both distribution-matched codebooks and hardware-native NVFP4, while commit-time KV-cache quantization extends the approach to block-diffusion models. Across full-sequence and block-diffusion dLLMs, \soloq achieves robust W4A4 quantization and remains competitive with or outperforms existing calibration-based baselines across a broad range of language benchmarks. These results establish predictable representation shaping as a general
foundation for efficient low-bit dLLM inference. Limitations are discussed in
Appendix~\ref{limit}.

\section*{Acknowledgments}
\vspace{-2mm}
This work was supported in part by CoCoSys, a JUMP2.0 center sponsored by DARPA and SRC, the National Science Foundation (CAREER Award, Grant \#2312366, Grant \#2318152), the DARPA Young Faculty Award, the DoE MMICC center SEA-CROGS (Award \#DE-SC0023198), and the Global Industrial Technology Cooperation Center (GITCC) program.

\newpage




\bibliography{iclr2027_conference}

@article{nie2026large,
  title={Large language diffusion models},
  author={Nie, Shen and Zhu, Fengqi and You, Zebin and Zhang, Xiaolu and Ou, Jingyang and Hu, Jun and Zhou, Jun and Lin, Yankai and Wen, Ji-Rong and Li, Chongxuan},
  journal={Advances in Neural Information Processing Systems},
  volume={38},
  pages={50608--50646},
  year={2026}
}

@article{ye2025dream,
  title={Dream 7B: Diffusion Large Language Models},
  author={Ye, Jiacheng and Xie, Zhihui and Zheng, Lin and Gao, Jiahui and Wu, Zirui and Jiang, Xin and Li, Zhenguo and Kong, Lingpeng},
  journal={arXiv preprint arXiv:2508.15487},
  year={2025}
}

@inproceedings{wu2026fast,
  title={Fast-dllm v2: Efficient block-diffusion llm},
  author={Wu, Chengyue and Zhang, Hao and Xue, Shuchen and Diao, Shizhe and Fu, Yonggan and Liu, Zhijian and Molchanov, Pavlo and Luo, Ping and Han, Song and Xie, Enze},
  booktitle={International Conference on Learning Representations},
  volume={2026},
  pages={128353--128370},
  year={2026}
}

@article{fu2026nemotron,
  title={Nemotron-Labs-Diffusion: A Tri-Mode Language Model Unifying Autoregressive, Diffusion, and Self-Speculation Decoding},
  author={Fu, Yonggan and Whalen, Lexington and Garg, Abhinav and Wu, Chengyue and Khadkevich, Maksim and Oswald, Nicolai and Xie, Enze and Egert, Daniel and Sreenivas, Sharath Turuvekere and Diao, Shizhe and others},
  journal={arXiv preprint arXiv:2607.05722},
  year={2026}
}

@article{brown2020language,
  title={Language models are few-shot learners},
  author={Brown, Tom and Mann, Benjamin and Ryder, Nick and Subbiah, Melanie and Kaplan, Jared D and Dhariwal, Prafulla and Neelakantan, Arvind and Shyam, Pranav and Sastry, Girish and Askell, Amanda and others},
  journal={Advances in neural information processing systems},
  volume={33},
  pages={1877--1901},
  year={2020}
}

@article{touvron2023llama,
  title={Llama: Open and efficient foundation language models},
  author={Touvron, Hugo and Lavril, Thibaut and Izacard, Gautier and Martinet, Xavier and Lachaux, Marie-Anne and Lacroix, Timoth{\'e}e and Rozi{\`e}re, Baptiste and Goyal, Naman and Hambro, Eric and Azhar, Faisal and others},
  journal={arXiv preprint arXiv:2302.13971},
  year={2023}
}

@inproceedings{peebles2023scalable,
  title={Scalable diffusion models with transformers},
  author={Peebles, William and Xie, Saining},
  booktitle={2023 IEEE/CVF International Conference on Computer Vision (ICCV)},
  pages={4172--4182},
  year={2023},
  organization={IEEE}
}

@inproceedings{bao2023all,
  title={All are worth words: A vit backbone for diffusion models},
  author={Bao, Fan and Nie, Shen and Xue, Kaiwen and Cao, Yue and Li, Chongxuan and Su, Hang and Zhu, Jun},
  booktitle={2023 IEEE/CVF Conference on Computer Vision and Pattern Recognition (CVPR)},
  pages={22669--22679},
  year={2023},
  organization={IEEE}
}

@article{achiam2023gpt,
  title={Gpt-4 technical report},
  author={Achiam, Josh and Adler, Steven and Agarwal, Sandhini and Ahmad, Lama and Akkaya, Ilge and Aleman, Florencia Leoni and Almeida, Diogo and Altenschmidt, Janko and Altman, Sam and Anadkat, Shyamal and others},
  journal={arXiv preprint arXiv:2303.08774},
  year={2023}
}

@article{team2023gemini,
  title={Gemini: a family of highly capable multimodal models},
  author={Team, Gemini and Anil, Rohan and Borgeaud, Sebastian and Alayrac, Jean-Baptiste and Yu, Jiahui and Soricut, Radu and Schalkwyk, Johan and Dai, Andrew M and Hauth, Anja and Millican, Katie and others},
  journal={arXiv preprint arXiv:2312.11805},
  year={2023}
}

@inproceedings{esser2024scaling,
  title={Scaling rectified flow transformers for high-resolution image synthesis},
  author={Esser, Patrick and Kulal, Sumith and Blattmann, Andreas and Entezari, Rahim and M{\"u}ller, Jonas and Saini, Harry and Levi, Yam and Lorenz, Dominik and Sauer, Axel and Boesel, Frederic and others},
  booktitle={Forty-first international conference on machine learning},
  year={2024}
}

@article{wu2025qwen,
  title={Qwen-image technical report},
  author={Wu, Chenfei and Li, Jiahao and Zhou, Jingren and Lin, Junyang and Gao, Kaiyuan and Yan, Kun and Yin, Sheng-ming and Bai, Shuai and Xu, Xiao and Chen, Yilei and others},
  journal={arXiv preprint arXiv:2508.02324},
  year={2025}
}

@article{kong2024hunyuanvideo,
  title={Hunyuanvideo: A systematic framework for large video generative models},
  author={Kong, Weijie and Tian, Qi and Zhang, Zijian and Min, Rox and Dai, Zuozhuo and Zhou, Jin and Xiong, Jiangfeng and Li, Xin and Wu, Bo and Zhang, Jianwei and others},
  journal={arXiv preprint arXiv:2412.03603},
  year={2024}
}

@inproceedings{yang2025cogvideox,
  title={Cogvideox: Text-to-video diffusion models with an expert transformer},
  author={Yang, Zhuoyi and Teng, Jiayan and Zheng, Wendi and Ding, Ming and Huang, Shiyu and Xu, Jiazheng and Yang, Yuanming and Hong, Wenyi and Zhang, Xiaohan and Feng, Guanyu and others},
  booktitle={International Conference on Learning Representations},
  volume={2025},
  pages={83048--83077},
  year={2025}
}

@article{team2026diffusiongemma,
  title={DiffusionGemma Technical Report},
  author={Team, DiffusionGemma and Ta{\"\i}ga, Adrien Ali and Assiene, James and Calandriello, Daniele and Chaabouni, Rahma and Gante, Jo{\~a}o and von Glehn, Tamara and Keating, Nate and Knutsen, Chris and Kukla, Martin and others},
  journal={arXiv preprint arXiv:2608.00146},
  year={2026}
}

@inproceedings{zhang2026quant,
  title={Quant-dllm: Post-training extreme low-bit quantization for diffusion large language models},
  author={Zhang, Tianao and Li, Zhiteng and Yan, Xianglong and Qin, Haotong and Guo, Yong and Zhang, Yulun},
  booktitle={International Conference on Learning Representations},
  volume={2026},
  pages={79492--79503},
  year={2026}
}

@article{xu2025dllmquant,
  title={Dllmquant: Quantizing diffusion-based large language models},
  author={Xu, Chen and Yang, Dawei},
  journal={arXiv preprint arXiv:2508.14090},
  year={2025}
}

@article{yan2026star,
  title={STaR-Quant: State-Time Consistent Post-Training Quantization for Diffusion Large Language Models},
  author={Yan, Xin and Wang, Aqiang and Wan, Zhenglin and Yu, Xingrui and Tsang, Ivor},
  journal={arXiv preprint arXiv:2606.04945},
  year={2026}
}

@article{huang2026fair,
  title={FAIR-Calib: Frontier-Aware Instability-Reweighted Calibration for Post-Training Quantization of Diffusion Large Language Models},
  author={Huang, Haoyu and Yang, Linlin and Xu, Sheng and Liu, Boyu and Guo, Guodong and Fu, Zhongqian and Zhou, Hang and Zhang, Baochang},
  journal={arXiv preprint arXiv:2606.06547},
  year={2026}
}

@article{frantar2022gptq,
  title={Gptq: Accurate post-training quantization for generative pre-trained transformers},
  author={Frantar, Elias and Ashkboos, Saleh and Hoefler, Torsten and Alistarh, Dan},
  journal={arXiv preprint arXiv:2210.17323},
  year={2022}
}

@article{li2025gptaq,
  title={Gptaq: Efficient finetuning-free quantization for asymmetric calibration},
  author={Li, Yuhang and Yin, Ruokai and Lee, Donghyun and Xiao, Shiting and Panda, Priyadarshini},
  journal={arXiv preprint arXiv:2504.02692},
  year={2025}
}

@article{lin2024awq,
  title={Awq: Activation-aware weight quantization for on-device llm compression and acceleration},
  author={Lin, Ji and Tang, Jiaming and Tang, Haotian and Yang, Shang and Chen, Wei-Ming and Wang, Wei-Chen and Xiao, Guangxuan and Dang, Xingyu and Gan, Chuang and Han, Song},
  journal={Proceedings of machine learning and systems},
  volume={6},
  pages={87--100},
  year={2024}
}

@article{lee2026kronq,
  title={KronQ: LLM Quantization via Kronecker-Factored Hessian},
  author={Lee, Donghyun and Li, Yuhang and Yin, Ruokai and Panda, Priyadarshini},
  journal={arXiv preprint arXiv:2607.07964},
  year={2026}
}

@inproceedings{xiao2023smoothquant,
  title={Smoothquant: Accurate and efficient post-training quantization for large language models},
  author={Xiao, Guangxuan and Lin, Ji and Seznec, Mickael and Wu, Hao and Demouth, Julien and Han, Song},
  booktitle={International conference on machine learning},
  pages={38087--38099},
  year={2023},
  organization={PMLR}
}

@article{ashkboos2024quarot,
  title={Quarot: Outlier-free 4-bit inference in rotated llms},
  author={Ashkboos, Saleh and Mohtashami, Amirkeivan and Croci, Maximilian L and Li, Bo and Cameron, Pashmina and Jaggi, Martin and Alistarh, Dan and Hoefler, Torsten and Hensman, James},
  journal={Advances in Neural Information Processing Systems},
  volume={37},
  pages={100213--100240},
  year={2024}
}

@article{chee2023quip,
  title={Quip: 2-bit quantization of large language models with guarantees},
  author={Chee, Jerry and Cai, Yaohui and Kuleshov, Volodymyr and De Sa, Christopher M},
  journal={Advances in neural information processing systems},
  volume={36},
  pages={4396--4429},
  year={2023}
}

@inproceedings{liu2025spinquant,
  title={Spinquant: Llm quantization with learned rotations},
  author={Liu, Zechun and Zhao, Changsheng and Fedorov, Igor and Soran, Bilge and Choudhary, Dhruv and Krishnamoorthi, Raghuraman and Chandra, Vikas and Tian, Yuandong and Blankevoort, Tijmen},
  booktitle={International Conference on Learning Representations},
  volume={2025},
  pages={92009--92032},
  year={2025}
}

@article{hu2025ostquant,
  title={Ostquant: Refining large language model quantization with orthogonal and scaling transformations for better distribution fitting},
  author={Hu, Xing and Cheng, Yuan and Yang, Dawei and Xu, Zukang and Yuan, Zhihang and Yu, Jiangyong and Xu, Chen and Jiang, Zhe and Zhou, Sifan},
  journal={arXiv preprint arXiv:2501.13987},
  year={2025}
}

@inproceedings{wei2023outlier,
  title={Outlier suppression+: Accurate quantization of large language models by equivalent and effective shifting and scaling},
  author={Wei, Xiuying and Zhang, Yunchen and Li, Yuhang and Zhang, Xiangguo and Gong, Ruihao and Guo, Jinyang and Liu, Xianglong},
  booktitle={Proceedings of the 2023 Conference on Empirical Methods in Natural Language Processing},
  pages={1648--1665},
  year={2023}
}

@inproceedings{shao2024omniquant,
  title={Omniquant: Omnidirectionally calibrated quantization for large language models},
  author={Shao, Wenqi and Chen, Mengzhao and Zhang, Zhaoyang and Xu, Peng and Zhao, Lirui and Li, Zhiqian and Zhang, Kaipeng and Peng, Gao and Qiao, Yu and Luo, Ping},
  booktitle={International Conference on Learning Representations},
  volume={2024},
  pages={45472--45496},
  year={2024}
}

@article{austin2021structured,
  title={Structured denoising diffusion models in discrete state-spaces},
  author={Austin, Jacob and Johnson, Daniel D and Ho, Jonathan and Tarlow, Daniel and Van Den Berg, Rianne},
  journal={Advances in neural information processing systems},
  volume={34},
  pages={17981--17993},
  year={2021}
}

@article{li2022diffusion,
  title={Diffusion-lm improves controllable text generation},
  author={Li, Xiang and Thickstun, John and Gulrajani, Ishaan and Liang, Percy S and Hashimoto, Tatsunori B},
  journal={Advances in neural information processing systems},
  volume={35},
  pages={4328--4343},
  year={2022}
}

@inproceedings{he2023diffusionbert,
  title={Diffusionbert: Improving generative masked language models with diffusion models},
  author={He, Zhengfu and Sun, Tianxiang and Tang, Qiong and Wang, Kuanning and Huang, Xuan-Jing and Qiu, Xipeng},
  booktitle={Proceedings of the 61st annual meeting of the association for computational linguistics (volume 1: Long papers)},
  pages={4521--4534},
  year={2023}
}

@inproceedings{zhu2026llada,
  title={Llada 1.5: Variance-reduced preference optimization for large language diffusion models},
  author={Zhu, Fengqi and Wang, Rongzhen and Nie, Shen and Zhang, Xiaolu and Wu, Chunwei and Zhou, Jun and Lin, Yankai and Wen, Ji-Rong and Li, Chongxuan},
  booktitle={Proceedings of the 64th Annual Meeting of the Association for Computational Linguistics (Volume 1: Long Papers)},
  pages={11425--11460},
  year={2026}
}

@misc{bie2026llada21speedingtextdiffusion,
      title={LLaDA2.1: Speeding Up Text Diffusion via Token Editing}, 
      author={Tiwei Bie and Maosong Cao and Xiang Cao and Bingsen Chen and Fuyuan Chen and Kun Chen and Lun Du and Daozhuo Feng and Haibo Feng and Mingliang Gong and Zhuocheng Gong and Yanmei Gu and Jian Guan and Kaiyuan Guan and Hongliang He and Zenan Huang and Juyong Jiang and Zhonghui Jiang and Zhenzhong Lan and Chengxi Li and Jianguo Li and Zehuan Li and Huabin Liu and Lin Liu and Guoshan Lu and Yuan Lu and Yuxin Ma and Xingyu Mou and Zhenxuan Pan and Kaida Qiu and Yuji Ren and Jianfeng Tan and Yiding Tian and Zian Wang and Lanning Wei and Tao Wu and Yipeng Xing and Wentao Ye and Liangyu Zha and Tianze Zhang and Xiaolu Zhang and Junbo Zhao and Da Zheng and Hao Zhong and Wanli Zhong and Jun Zhou and Junlin Zhou and Liwang Zhu and Muzhi Zhu and Yihong Zhuang},
      year={2026},
      eprint={2602.08676},
      archivePrefix={arXiv},
      primaryClass={cs.LG},
      url={https://arxiv.org/abs/2602.08676}, 
}

@misc{bie2025llada20scalingdiffusionlanguage,
      title={LLaDA2.0: Scaling Up Diffusion Language Models to 100B}, 
      author={Tiwei Bie and Maosong Cao and Kun Chen and Lun Du and Mingliang Gong and Zhuochen Gong and Yanmei Gu and Jiaqi Hu and Zenan Huang and Zhenzhong Lan and Chengxi Li and Chongxuan Li and Jianguo Li and Zehuan Li and Huabin Liu and Ling Liu and Guoshan Lu and Xiaocheng Lu and Yuxin Ma and Jianfeng Tan and Lanning Wei and Ji-Rong Wen and Yipeng Xing and Xiaolu Zhang and Junbo Zhao and Da Zheng and Jun Zhou and Junlin Zhou and Zhanchao Zhou and Liwang Zhu and Yihong Zhuang},
      year={2025},
      eprint={2512.15745},
      archivePrefix={arXiv},
      primaryClass={cs.LG},
      url={https://arxiv.org/abs/2512.15745}, 
}

@inproceedings{arriola2025block,
  title={Block diffusion: Interpolating between autoregressive and diffusion language models},
  author={Arriola, Marianne and Gokaslan, Aaron and Chiu, Justin and Yang, Zhihan and Qi, Zhixuan and Han, Jiaqi and Sahoo, Subham and Kuleshov, Volodymyr},
  booktitle={International Conference on Learning Representations},
  volume={2025},
  pages={50726--50753},
  year={2025}
}

@inproceedings{wu2026fast1,
  title={Fast-dllm: Training-free acceleration of diffusion llm by enabling kv cache and parallel decoding},
  author={Wu, Chengyue and Zhang, Hao and Xue, Shuchen and Liu, Zhijian and Diao, Shizhe and Zhu, Ligeng and Luo, Ping and Han, Song and Xie, Enze},
  booktitle={International Conference on Learning Representations},
  volume={2026},
  pages={57027--57051},
  year={2026}
}

@article{reda2026nemotron,
  title={Nemotron-TwoTower: Diffusion Language Modeling with Pretrained Autoregressive Context},
  author={Reda, Fitsum and Kamalu, John and Waleffe, Roger and Patwary, Mostofa and Shoeybi, Mohammad and Catanzaro, Bryan},
  journal={arXiv preprint arXiv:2606.26493},
  year={2026}
}

@article{tseng2024quip,
  title={Quip\#: Even better llm quantization with hadamard incoherence and lattice codebooks},
  author={Tseng, Albert and Chee, Jerry and Sun, Qingyao and Kuleshov, Volodymyr and De Sa, Christopher},
  journal={Proceedings of machine learning research},
  volume={235},
  pages={48630},
  year={2024}
}

@inproceedings{zandieh2026turboquant,
  title={Turboquant: Online vector quantization with near-optimal distortion rate},
  author={Zandieh, Amir and Daliri, Majid and Hadian, Majid and Mirrokni, Vahab},
  booktitle={International Conference on Learning Representations},
  volume={2026},
  pages={56418--56439},
  year={2026}
}

@article{han2025polarquant,
  title={Polarquant: Quantizing kv caches with polar transformation},
  author={Han, Insu and Kacham, Praneeth and Karbasi, Amin and Mirrokni, Vahab and Zandieh, Amir},
  journal={arXiv preprint arXiv:2502.02617},
  year={2025}
}

@article{lee2026orbitquant,
  title={OrbitQuant: Data-Agnostic Quantization for Image and Video Diffusion Transformers},
  author={Lee, Donghyun and Chavan, Jitesh and Nguyen, Duy and Huang, Sam and Jiang, Liming and Panda, Priyadarshini and Mertens, Timo and Shukla, Saurabh},
  journal={arXiv preprint arXiv:2607.02461},
  year={2026}
}

@article{huang2025convrot,
  title={ConvRot: Rotation-Based Plug-and-Play 4-bit Quantization for Diffusion Transformers},
  author={Huang, Feice and Han, Zuliang and Zhou, Xing and Chen, Yihuang and Zhu, Lifei and Wang, Haoqian},
  journal={arXiv preprint arXiv:2512.03673},
  year={2025}
}

@inproceedings{li2026dvd,
  title={Dvd-quant: data-free video diffusion transformers quantization},
  author={Li, Zhiteng and Li, Hanxuan and Wu, Junyi and Liu, Kai and Qin, Haotong and Kong, Linghe and Chen, Guihai and Zhang, Yulun and Yang, Xiaokang},
  booktitle={International Conference on Learning Representations},
  volume={2026},
  pages={35463--35475},
  year={2026}
}

@article{vargaftik2021drive,
  title={Drive: One-bit distributed mean estimation},
  author={Vargaftik, Shay and Ben-Basat, Ran and Portnoy, Amit and Mendelson, Gal and Ben-Itzhak, Yaniv and Mitzenmacher, Michael},
  journal={Advances in Neural Information Processing Systems},
  volume={34},
  pages={362--377},
  year={2021}
}

@inproceedings{vargaftik2022eden,
  title={Eden: Communication-efficient and robust distributed mean estimation for federated learning},
  author={Vargaftik, Shay and Basat, Ran Ben and Portnoy, Amit and Mendelson, Gal and Itzhak, Yaniv Ben and Mitzenmacher, Michael},
  booktitle={International Conference on Machine Learning},
  pages={21984--22014},
  year={2022},
  organization={PMLR}
}

@article{gao2024rabitq,
  title={Rabitq: Quantizing high-dimensional vectors with a theoretical error bound for approximate nearest neighbor search},
  author={Gao, Jianyang and Long, Cheng},
  journal={Proceedings of the ACM on Management of Data},
  volume={2},
  number={3},
  pages={1--27},
  year={2024},
  publisher={ACM New York, NY, USA}
}

@article{liu2024kivi,
  title={Kivi: A tuning-free asymmetric 2bit quantization for kv cache},
  author={Liu, Zirui and Yuan, Jiayi and Jin, Hongye and Zhong, Shaochen and Xu, Zhaozhuo and Braverman, Vladimir and Chen, Beidi and Hu, Xia},
  journal={arXiv preprint arXiv:2402.02750},
  year={2024}
}

@article{su2026oscar,
  title={OScaR: The Occam's Razor for Extreme KV Cache Quantization in LLMs and Beyond},
  author={Su, Zunhai and Yang, Rui and Zhang, Chao and Liu, Yaxiu and Zhang, Yifan and Wu, Wei and Xiong, Jing and Du, Dayou and Zhuang, Xialie and Qian, Yulei and others},
  journal={arXiv preprint arXiv:2605.19660},
  year={2026}
}

@inproceedings{lin2022truthfulqa,
  title={Truthfulqa: Measuring how models mimic human falsehoods},
  author={Lin, Stephanie and Hilton, Jacob and Evans, Owain},
  booktitle={Proceedings of the 60th annual meeting of the association for computational linguistics (volume 1: long papers)},
  pages={3214--3252},
  year={2022}
}

@article{clark2018think,
  title={Think you have solved question answering? try arc, the ai2 reasoning challenge},
  author={Clark, Peter and Cowhey, Isaac and Etzioni, Oren and Khot, Tushar and Sabharwal, Ashish and Schoenick, Carissa and Tafjord, Oyvind},
  journal={arXiv preprint arXiv:1803.05457},
  year={2018}
}

@inproceedings{zellers2019hellaswag,
  title={Hellaswag: Can a machine really finish your sentence?},
  author={Zellers, Rowan and Holtzman, Ari and Bisk, Yonatan and Farhadi, Ali and Choi, Yejin},
  booktitle={Proceedings of the 57th annual meeting of the association for computational linguistics},
  pages={4791--4800},
  year={2019}
}

@article{sakaguchi2021winogrande,
  title={Winogrande: An adversarial winograd schema challenge at scale},
  author={Sakaguchi, Keisuke and Bras, Ronan Le and Bhagavatula, Chandra and Choi, Yejin},
  journal={Communications of the ACM},
  volume={64},
  number={9},
  pages={99--106},
  year={2021},
  publisher={ACM New York, NY, USA}
}

@inproceedings{bisk2020piqa,
  title={Piqa: Reasoning about physical commonsense in natural language},
  author={Bisk, Yonatan and Zellers, Rowan and Gao, Jianfeng and Choi, Yejin and others},
  booktitle={Proceedings of the AAAI conference on artificial intelligence},
  volume={34},
  number={05},
  pages={7432--7439},
  year={2020}
}

@article{huang2023c,
  title={C-eval: A multi-level multi-discipline chinese evaluation suite for foundation models},
  author={Huang, Yuzhen and Bai, Yuzhuo and Zhu, Zhihao and Zhang, Junlei and Zhang, Jinghan and Su, Tangjun and Liu, Junteng and Lv, Chuancheng and Zhang, Yikai and Fu, Yao and others},
  journal={Advances in neural information processing systems},
  volume={36},
  pages={62991--63010},
  year={2023}
}

@article{chen2021evaluating,
  title={Evaluating large language models trained on code},
  author={Chen, Mark and Tworek, Jerry and Jun, Heewoo and Yuan, Qiming and Pinto, Henrique Ponde De Oliveira and Kaplan, Jared and Edwards, Harri and Burda, Yuri and Joseph, Nicholas and Brockman, Greg and others},
  journal={arXiv preprint arXiv:2107.03374},
  year={2021}
}

@article{cobbe2021training,
  title={Training verifiers to solve math word problems},
  author={Cobbe, Karl and Kosaraju, Vineet and Bavarian, Mohammad and Chen, Mark and Jun, Heewoo and Kaiser, Lukasz and Plappert, Matthias and Tworek, Jerry and Hilton, Jacob and Nakano, Reiichiro and others},
  journal={arXiv preprint arXiv:2110.14168},
  year={2021}
}

@article{austin2021program,
  title={Program synthesis with large language models},
  author={Austin, Jacob and Odena, Augustus and Nye, Maxwell and Bosma, Maarten and Michalewski, Henryk and Dohan, David and Jiang, Ellen and Cai, Carrie and Terry, Michael and Le, Quoc and others},
  journal={arXiv preprint arXiv:2108.07732},
  year={2021}
}

@article{liu2023your,
  title={Is your code generated by chatgpt really correct? rigorous evaluation of large language models for code generation},
  author={Liu, Jiawei and Xia, Chunqiu Steven and Wang, Yuyao and Zhang, Lingming},
  journal={Advances in neural information processing systems},
  volume={36},
  pages={21558--21572},
  year={2023}
}

@article{hendrycks2021measuring,
  title={Measuring mathematical problem solving with the math dataset},
  author={Hendrycks, Dan and Burns, Collin and Kadavath, Saurav and Arora, Akul and Basart, Steven and Tang, Eric and Song, Dawn and Steinhardt, Jacob},
  journal={arXiv preprint arXiv:2103.03874},
  year={2021}
}

@article{zhou2023instruction,
  title={Instruction-following evaluation for large language models},
  author={Zhou, Jeffrey and Lu, Tianjian and Mishra, Swaroop and Brahma, Siddhartha and Basu, Sujoy and Luan, Yi and Zhou, Denny and Hou, Le},
  journal={arXiv preprint arXiv:2311.07911},
  year={2023}
}

@article{hendrycks2020measuring,
  title={Measuring massive multitask language understanding},
  author={Hendrycks, Dan and Burns, Collin and Basart, Steven and Zou, Andy and Mazeika, Mantas and Song, Dawn and Steinhardt, Jacob},
  journal={arXiv preprint arXiv:2009.03300},
  year={2020}
}

@article{rein2023gpqa,
  title={Gpqa: A graduate-level google-proof q\&a benchmark},
  author={Rein, David and Hou, Betty Li and Stickland, Asa Cooper and Petty, Jackson and Pang, Richard Yuanzhe and Dirani, Julien and Michael, Julian and Bowman, Samuel R},
  journal={arXiv preprint arXiv:2311.12022},
  year={2023}
}

@misc{eval-harness,
  author       = {Gao, Leo and Tow, Jonathan and Abbasi, Baber and Biderman, Stella and Black, Sid and DiPofi, Anthony and Foster, Charles and Golding, Laurence and Hsu, Jeffrey and Le Noac'h, Alain and Li, Haonan and McDonell, Kyle and Muennighoff, Niklas and Ociepa, Chris and Phang, Jason and Reynolds, Laria and Schoelkopf, Hailey and Skowron, Aviya and Sutawika, Lintang and Tang, Eric and Thite, Anish and Wang, Ben and Wang, Kevin and Zou, Andy},
  title        = {The Language Model Evaluation Harness},
  month        = 07,
  year         = 2024,
  publisher    = {Zenodo},
  version      = {v0.4.3},
  doi          = {10.5281/zenodo.12608602},
  url          = {https://zenodo.org/records/12608602}
}

@misc{2023opencompass,
    title={OpenCompass: A Universal Evaluation Platform for Foundation Models},
    author={OpenCompass Contributors},
    howpublished = {\url{https://github.com/open-compass/opencompass}},
    year={2023}
}

@inproceedings{xiao2024efficient,
  title={Efficient streaming language models with attention sinks},
  author={Xiao, Guangxuan and Tian, Yuandong and Chen, Beidi and Han, Song and Lewis, Mike},
  booktitle={International Conference on Learning Representations},
  volume={2024},
  pages={21875--21895},
  year={2024}
}

@inproceedings{rulli2026attention,
  title={Attention sinks in diffusion language models},
  author={Rulli, Maximo Eduardo and Petruzzi, Simone and Michielon, Edoardo and Silvestri, Fabrizio and Scardapane, Simone and Devoto, Alessio},
  booktitle={Findings of the Association for Computational Linguistics: ACL 2026},
  pages={43244--43256},
  year={2026}
}
\bibliographystyle{iclr2027_conference}

\newpage
\appendix
\section{Proof of Proposition~\ref{prop:krpbh_variance}}
\label{app:proof_krpbh}

Let $y=\mathbf{P}x$ and partition it into $k$ blocks
$y^{(1)},\ldots,y^{(k)}\in\mathbb{R}^{h}$.
Define the energy of the $j$-th block as
\begin{equation}
M_j
=
\left\|y^{(j)}\right\|_2^2,
\qquad
\sum_{j=1}^{k} M_j = 1,
\label{eq:app_block_mass}
\end{equation}
where the equality follows from $\|x\|_2=1$ and the orthogonality of
the permutation matrix $\mathbf{P}$.

Before the cross-block mixing, the output of the $j$-th RPBH block is
\begin{equation}
z^{(j)}
=
\mathbf{H}_h \mathbf{D}_j y^{(j)}.
\label{eq:app_local_rpbh}
\end{equation}
For its $\ell$-th coordinate,
\begin{equation}
z_{\ell}^{(j)}
=
\sum_{m=1}^{h}
H_{\ell m}
D_{j,mm}
y_m^{(j)}.
\label{eq:app_local_coordinate}
\end{equation}
Since the diagonal entries of $\mathbf{D}_j$ are independent
Rademacher variables with zero mean and unit variance, and the
normalized Walsh--Hadamard matrix satisfies
$H_{\ell m}^2=1/h$, we obtain
\begin{equation}
\mathbb{E}_{\mathbf{D}_j}
\left[z_{\ell}^{(j)}\right]
=0,
\qquad
\operatorname{Var}_{\mathbf{D}_j}
\left(z_{\ell}^{(j)}\right)
=
\frac{M_j}{h}.
\label{eq:app_local_variance}
\end{equation}

K-RPBH subsequently applies
$\mathbf{Q}_k\otimes\mathbf{I}_h$, which mixes the same
within-block coordinate across the $k$ blocks. Thus, the $\ell$-th
coordinate of output block $r$ is
\begin{equation}
\hat{z}_{\ell}^{(r)}
=
\sum_{j=1}^{k}
Q_{rj} z_{\ell}^{(j)}.
\label{eq:app_cross_block}
\end{equation}
Conditioned on $\mathbf{P}$ and $\mathbf{Q}_k$, the variables
$z_{\ell}^{(1)},\ldots,z_{\ell}^{(k)}$ are independent because the
sign matrices $\mathbf{D}_1,\ldots,\mathbf{D}_k$ are sampled
independently. Therefore,
\begin{align}
\operatorname{Var}_{\mathbf{D}}
\left(
\hat{z}_{\ell}^{(r)}
\mid
\mathbf{P},\mathbf{Q}_k
\right)
&=
\sum_{j=1}^{k}
Q_{rj}^{2}
\operatorname{Var}_{\mathbf{D}_j}
\left(z_{\ell}^{(j)}\right)
\nonumber\\
&=
\frac{1}{h}
\sum_{j=1}^{k}
Q_{rj}^{2} M_j,
\label{eq:app_cross_block_variance}
\end{align}
which proves Equation~\ref{eq:krpbh_var}.

Finally, each row of a Haar-random orthogonal matrix
$\mathbf{Q}_k$ is uniformly distributed on the unit sphere
$\mathbb{S}^{k-1}$, and hence
\begin{equation}
\mathbb{E}_{\mathbf{Q}_k}
\left[Q_{rj}^{2}\right]
=
\frac{1}{k}.
\label{eq:app_haar_second_moment}
\end{equation}
Taking expectation over $\mathbf{Q}_k$ in
Equation~\ref{eq:app_cross_block_variance} gives
\begin{align}
\mathbb{E}_{\mathbf{Q}_k}
\left[
\operatorname{Var}_{\mathbf{D}}
\left(
\hat{z}_{\ell}^{(r)}
\mid
\mathbf{P},\mathbf{Q}_k
\right)
\right]
&=
\frac{1}{h}
\sum_{j=1}^{k}
\mathbb{E}_{\mathbf{Q}_k}
\left[Q_{rj}^{2}\right] M_j
\nonumber\\
&=
\frac{1}{hk}
\sum_{j=1}^{k}M_j
=
\frac{1}{hk}
=
\frac{1}{d},
\label{eq:app_expected_variance}
\end{align}
where we use $\sum_j M_j=1$ and $d=kh$.
This proves Equation~\ref{eq:krpbh_expected_var}.
\hfill$\square$

\newpage

\section{Algorithm}
\label{Apix:algo}
For compactness, let
$\mathrm{SQ}^{W}_{\nu,b}(\cdot)$ and
$\mathrm{SQ}^{A}_{\nu,b}(\cdot)$ denote the \soloq weight and activation quantization
operators, respectively, for variant $\nu\in\{\mathrm{C},\mathrm{N}\}$ and bit-width $b$.
Each operator includes its respective scaling, quantization, and scale-only bias correction.
For committed keys, let $\mathrm{AsymPC}_{b}(\cdot)$ denote
$b$-bit per-channel asymmetric quantization along the token dimension.
Here, $\mathcal{D}$ denotes the set of rotation dimensions used by the target layers and KV values,
$N$ the number of tokens, $B$ the active block, and $d_v$ the value-vector dimension.

\begin{algorithm}[h]
\caption{\soloq offline weight and online activation/KV-cache quantization}
\label{alg:soloq}
\begin{algorithmic}[1]

\REQUIRE Transformer $\mathcal{T}$, target layers $\mathcal{L}$,
rotation dimensions $\mathcal{D}$, variant $\nu\in\{\mathrm{C},\mathrm{N}\}$,
bit-widths $(b_w,b_a,b_{\mathrm{kv}})$

\medskip
\STATE $\triangleright$ \textbf{Offline}
\FOR{$d\in\mathcal{D}$}
    \STATE $\Pi_d \leftarrow \mathrm{K\mbox{-}RPBH}(d)$
\ENDFOR

\FOR{each $\mathbf{W}\in\mathcal{L}$ with input dimension $d$}
    \STATE $\mathbf{W}' \leftarrow \mathbf{W}\Pi_d^\top$
    \STATE $\widehat{\mathbf{W}}'
    \leftarrow \mathrm{SQ}^{W}_{\nu,b_w}(\mathbf{W}')$
    \STATE Replace $\mathbf{W}$ with $\widehat{\mathbf{W}}'$ in $\mathcal{T}$
\ENDFOR

\medskip
\STATE $\triangleright$ \textbf{Online activation quantization}
for $\mathbf{X}\in\mathbb{R}^{N\times d}$
\STATE $\mathbf{X}' \leftarrow \mathbf{X}\Pi_d^\top$
\STATE $\widehat{\mathbf{X}}'
\leftarrow \mathrm{SQ}^{A}_{\nu,b_a}(\mathbf{X}')$

\medskip
\STATE $\triangleright$ \textbf{Commit-time KV quantization} for block-diffusion models
\STATE Keep the active KV block $(\mathbf{K}_B,\mathbf{V}_B)$ in full precision
\STATE On commit,
$\widehat{\mathbf{K}}_B
\leftarrow \mathrm{AsymPC}_{b_{\mathrm{kv}}}(\mathbf{K}_B)$
\STATE On commit,
$\widehat{\mathbf{V}}_B
\leftarrow \mathrm{SQ}^{A}_{\nu,b_{\mathrm{kv}}}
\!\left(\mathbf{V}_B\Pi_{d_v}^{\top}\right)$
\STATE Append $(\widehat{\mathbf{K}}_B,\widehat{\mathbf{V}}_B)$
to the persistent KV-cache

\RETURN Quantized $\mathcal{T}$ and persistent KV-cache

\end{algorithmic}
\end{algorithm}

\vspace{-5mm}
\section{Evaluation Details}
\label{app:eval_setup}
\vspace{-3mm}
We retain each model's official evaluation stack rather than enforcing a unified harness, since identically named benchmarks can differ in prompting, inference, and grading. Within each model, FP and quantized runs use identical prompts, few-shot settings, batch sizes, generation parameters, and graders. For full-sequence models, we evaluate TruthfulQA~\citep{lin2022truthfulqa}, ARC-Challenge~\citep{clark2018think}, HellaSwag~\citep{zellers2019hellaswag}, WinoGrande~\citep{sakaguchi2021winogrande}, PIQA~\citep{bisk2020piqa}, MMLU~\citep{hendrycks2020measuring}, C-Eval~\citep{huang2023c}, HumanEval~\citep{chen2021evaluating}, and GSM8K~\citep{cobbe2021training}. For block-diffusion models, Human-B/P and MBPP-B/P denote the Base and Plus variants of HumanEval~\citep{chen2021evaluating} and MBPP~\citep{austin2021program}, evaluated using EvalPlus~\citep{liu2023your}. We additionally evaluate GSM8K~\citep{cobbe2021training}, MATH~\citep{hendrycks2021measuring}, IFEval~\citep{zhou2023instruction}, MMLU~\citep{hendrycks2020measuring}, and GPQA~\citep{rein2023gpqa}.

\vspace{-2mm}
\subsection{LLaDA-8B-Base}
\vspace{-2mm}

For LLaDA-8B-Base, we follow the official LM Evaluation Harness setup~\citep{eval-harness}. TruthfulQA-MC2, ARC-Challenge, HellaSwag, PIQA, and WinoGrande use 128 Monte Carlo masking samples, while MMLU and C-EVAL use one. TruthfulQA uses CFG 2.0, ARC-Challenge/HellaSwag/PIQA use CFG 0.5, and the remaining tasks use CFG 0. WinoGrande, MMLU, and C-EVAL are 5-shot; the other likelihood tasks are zero-shot.
All likelihood tasks report plain accuracy. GSM8K follows the official 5-shot setup with generation length, diffusion steps, and block length of 1,024 and strict-match scoring. HumanEval is zero-shot with the same generation configuration and reports pass@1.
\vspace{-2mm}

\subsection{LLaDA-1.5}
\vspace{-2mm}

LLaDA-1.5 is evaluated with its official OpenCompass stack~\citep{2023opencompass} for ARC-Challenge, HellaSwag, MMLU, GSM8K, and HumanEval. ARC-Challenge uses zero-shot generation with length/steps/block size 512; HellaSwag and MMLU use generation length/steps/block size 3, with MMLU evaluated 5-shot. GSM8K uses four in-prompt demonstrations with 256/256/16, and HumanEval uses 512/512/32. All generation tasks use greedy decoding and low-confidence remasking.
ARC-Challenge and HellaSwag use first-option accuracy, MMLU reports the macro-average across 57 subjects, GSM8K uses answer extraction, and HumanEval reports pass@1. For compatibility with prior quantization work, TruthfulQA-MC2, WinoGrande, PIQA, and C-EVAL are additionally evaluated using the LLaDA-8B likelihood protocol; these columns are supplementary and not part of the official LLaDA-1.5 generation suite.
\vspace{-2mm}

\subsection{Dream-v0-Base-7B}
\vspace{-2mm}

Dream-v0-Base-7B is evaluated with its native evaluator, including the Monte Carlo ELBO estimator, autoregressive logit shift, and BOS token. All likelihood tasks use 128 Monte Carlo samples. ARC-Challenge, HellaSwag, PIQA, and TruthfulQA-MC2 are zero-shot, while WinoGrande, MMLU, and C-EVAL are 5-shot. We use batch size 32 for ARC-Challenge/HellaSwag/PIQA/WinoGrande/TruthfulQA, 16 for MMLU, and 1 for C-EVAL; the reduced MMLU batch size is applied consistently to FP and quantized runs.
ARC-Challenge, HellaSwag, and PIQA report normalized accuracy, while the remaining likelihood tasks report accuracy. GSM8K uses 8-shot CoT prompting with 256 generated tokens and 256 diffusion steps; HumanEval is zero-shot with 512 generated tokens and 512 diffusion steps. TruthfulQA and C-EVAL are used only for controlled FP-to-quantized comparisons.
\vspace{-2mm}

\subsection{Fast-dLLM v2}
\vspace{-2mm}

For Fast-dLLM v2, evaluation follows the official NVLabs adapter. GSM8K, Minerva MATH, and IFEval are zero-shot with batch size 32; MMLU is zero-shot and GPQA Main 5-shot with batch size 1. Generation uses block size 32, sub-block size 8, threshold 1, greedy decoding, and up to 2,048 new tokens. MMLU and GPQA are evaluated by likelihood rather than generation.
GSM8K uses flexible answer extraction, Minerva MATH uses symbolic-equivalence rescoring, and IFEval reports prompt-level strict accuracy. HumanEval and MBPP use EvalPlus with batch size 32, reporting both base and Plus pass@1 after code sanitization.
\vspace{-2mm}

\subsection{Nemotron-Labs-Diffusion-8B}
\vspace{-2mm}

Nemotron-Labs-Diffusion-8B is evaluated with the official inference implementation and NeMo-Skills stack. All tasks, including MMLU and GPQA, are evaluated through generation with block length 32, threshold 0.9, temperature 0, batch size 1, and up to 8,192 generated tokens. Thinking mode and EOS early stopping are disabled. Since the Nemotron paper does not specify block length or EOS policy for its main table, we treat this as a controlled operating point rather than an exact reconstruction.
For the cross-model table, we evaluate GSM8K, Hendrycks MATH, IFEval, MMLU, GPQA Main, HumanEval, and MBPP. Mathematical tasks use the NeMo-Skills answer evaluator, MMLU and GPQA use generation-based multiple-choice scoring, HumanEval and MBPP use EvalPlus, and IFEval is averaged over its four strict/loose instruction-following metrics. MATH-500 and GPQA Diamond are reported only in comparisons matching the Nemotron paper's benchmark configuration.

\section{Additional Analysis}

\subsection{Comparisons with FAIR-Calib}
\label{fair-calib}
FAIR-Calib~\citep{huang2026fair} is a calibration-based PTQ method for dLLMs that prioritizes fragile token states near the write frontier, where quantization errors can alter token decisions before they are committed.
It first probes a full-precision teacher to estimate a frontier-aware position prior, then performs off-policy layer-wise calibration with a reweighted hidden-state reconstruction loss.
Since FAIR-Calib uses an evaluation suite different from our main full-sequence setting, we report a separate W4A4 comparison on the five shared benchmarks: TruthfulQA, WinoGrande, PIQA, MMLU, and GSM8K.
As shown in Table~\ref{tab:faircalib_comparison}, \soloq achieves higher average accuracy on both LLaDA-Base and Dream-7B without calibration data.

\begin{table*}[h]
\caption{
Comparison with FAIR-Calib under W4A4 quantization on the five benchmarks shared by both evaluation settings.
Results for FP16, RTN, QuaRot, FlatQuant, and FAIR-Calib are taken from FAIR-Calib, while the remaining baselines and \soloq results follow our evaluation.
Values are task accuracy (\%) and their average.
\textbf{Bold} and \underline{underlined} indicate the best and second-best quantized results within each model.
}
\label{tab:faircalib_comparison}
\small
\centering
\resizebox{0.9\textwidth}{!}{%
\begin{tabular}{clccccccc}
\toprule
Model
& Method
& Calibration-Free
& Truth.
& Wino.
& PIQA
& MMLU
& GSM8K
& Avg. \\
\midrule

\multirow{9}{*}{LLaDA-Base}

& FP16
& -
& 45.30
& 73.64
& 74.84
& 65.80
& 68.92
& 65.70 \\
\cdashline{2-9}

& RTN
& \cmark
& 38.80
& 61.80
& 69.26
& 51.05
& 35.03
& 51.19 \\

& AWQ
& \xmark
& 40.87
& 66.88
& 69.43
& 51.22
& 36.88
& 53.06 \\

& QuaRot
& \xmark
& 43.35
& 72.22
& 73.61
& 62.04
& 57.39
& 61.72 \\

& FlatQuant
& \xmark
& 42.90
& 72.16
& 74.16
& 63.80
& 57.24
& 62.05 \\

& DLLMQuant+
& \xmark
& 41.53
& 67.87
& 70.12
& 51.72
& 40.66
& 54.38 \\

& DLLMQuant++
& \xmark
& 43.53
& 71.85
& 73.94
& 57.77
& 56.25
& 60.67 \\

& FAIR-Calib
& \xmark
& 43.91
& 72.93
& \textbf{74.92}
& \underline{64.11}
& 64.75
& 64.12 \\

& \cellcolor{gray!25}\soloq-C
& \cellcolor{gray!25}\cmark
& \cellcolor{gray!25}\underline{45.22}
& \cellcolor{gray!25}\underline{74.51}
& \cellcolor{gray!25}74.16
& \cellcolor{gray!25}\underline{64.11}
& \cellcolor{gray!25}\textbf{67.78}
& \cellcolor{gray!25}\underline{65.16} \\

& \cellcolor{gray!25}\soloq-N
& \cellcolor{gray!25}\cmark
& \cellcolor{gray!25}\textbf{45.33}
& \cellcolor{gray!25}\textbf{74.59}
& \cellcolor{gray!25}\underline{74.48}
& \cellcolor{gray!25}\textbf{64.55}
& \cellcolor{gray!25}\underline{67.48}
& \cellcolor{gray!25}\textbf{65.29} \\

\midrule

\multirow{9}{*}{Dream-7B}

& FP16
& -
& 44.17
& 73.32
& 75.41
& 71.36
& 76.19
& 68.09 \\
\cdashline{2-9}

& RTN
& \cmark
& 41.42
& 55.64
& 61.70
& 36.94
& 16.83
& 42.51 \\

& AWQ
& \xmark
& 43.66
& 67.13
& 68.96
& 55.50
& 48.98
& 56.85 \\

& QuaRot
& \xmark
& 40.40
& 63.46
& 72.52
& 59.74
& 49.51
& 57.13 \\

& FlatQuant
& \xmark
& 43.23
& 65.98
& 71.65
& 64.96
& 60.20
& 61.20 \\

& DLLMQuant+
& \xmark
& \underline{44.14}
& 68.66
& 69.53
& 57.63
& 51.14
& 58.22 \\

& DLLMQuant++
& \xmark
& \textbf{47.86}
& \underline{71.54}
& 72.09
& \textbf{69.50}
& 66.17
& 65.43 \\

& FAIR-Calib
& \xmark
& 43.40
& 69.53
& 73.07
& 66.91
& 66.64
& 63.91 \\

& \cellcolor{gray!25}\soloq-C
& \cellcolor{gray!25}\cmark
& \cellcolor{gray!25}41.99
& \cellcolor{gray!25}71.35
& \cellcolor{gray!25}\textbf{74.76}
& \cellcolor{gray!25}68.86
& \cellcolor{gray!25}\textbf{72.86}
& \cellcolor{gray!25}\textbf{65.96} \\

& \cellcolor{gray!25}\soloq-N
& \cellcolor{gray!25}\cmark
& \cellcolor{gray!25}42.95
& \cellcolor{gray!25}\textbf{72.22}
& \cellcolor{gray!25}\underline{74.54}
& \cellcolor{gray!25}\underline{68.97}
& \cellcolor{gray!25}\underline{70.36}
& \cellcolor{gray!25}\underline{65.81} \\
\bottomrule
\end{tabular}%
}
\end{table*}

\subsection{Uniform INT4 Quantization}

To evaluate whether \soloq depends on specialized non-uniform quantizers, we additionally replace the codebook and NVFP4 quantizers with standard uniform INT4 quantization while keeping the K-RPBH rotation and the remaining pipeline unchanged. We denote this variant as \soloq-I. Tables~\ref{tab:int4_full_dllm} and~\ref{tab:int4_fastdllm} compare \soloq-I with \soloq-C, \soloq-N, and existing W4A4 baselines.
\soloq-I remains competitive across all three architectures, achieving average accuracies of 56.98 on LLaDA-Base, 67.44 on LLaDA-1.5, and 55.32 on Fast-dLLM v2. The degradation relative to \soloq-C/N is modest, indicating that the predictable representation produced by K-RPBH also supports conventional uniform INT4 quantization without calibration.

\begin{table*}[h]
\caption{
Comparison with uniform INT4 quantization on full-sequence diffusion language models.
\soloq-C, \soloq-N, and \soloq-I denote the codebook, NVFP4, and uniform INT4 variants, respectively.
All \soloq variants use the same unfolded K-RPBH topology.
Values are task accuracy (\%) and their average.
}
\label{tab:int4_full_dllm}
\small
\centering
\resizebox{\textwidth}{!}{%
\begin{tabular}{clcccccccccc}
\toprule
Model & Method
& Truth.
& ARC-C
& Hella.
& Wino.
& PIQA
& MMLU
& C-EVAL
& Human.
& GSM8K
& Avg. \\
\midrule


\multirow{10}{*}{LLaDA-Base}
& FP16
& 47.45 & 44.03 & 54.09 & 74.82 & 74.81 & 64.15 & 69.99 & 31.71 & 69.52 & 58.95 \\
\cdashline{2-12}

& RTN
& 40.45 & 41.83 & 45.40 & 64.72 & 67.95 & 49.26 & 57.95 & 14.02 & 16.56 & 44.23 \\

& AWQ
& 40.87 & 42.92 & 46.14 & 66.88 & 69.43 & 51.22 & 58.43 & 20.10 & 36.88 & 48.09 \\

& QuaRot
& 42.53 & 44.20 & 49.76 & 69.85 & 70.75 & 55.96 & 56.32 & 25.33 & 44.57 & 51.03 \\

& DLLMQuant+
& 41.53 & 43.44 & 46.51 & 67.87 & 70.12 & 51.72 & 59.38 & 22.13 & 40.66 & 49.26 \\

& DLLMQuant++
& 43.53 & 44.18 & 51.00 & 71.85 & 73.94 & 57.77 & 61.22 & 28.92 & 56.25 & 54.29 \\

& STaR-Quant
& 49.12 & 44.23 & 52.75 & 72.92 & 73.85 & 62.95 & 64.56 & 35.98 & 57.29 & 57.07 \\

\cdashline{2-12}

& \cellcolor{gray!25}\soloq-C
& \cellcolor{gray!25}45.22
& \cellcolor{gray!25}43.52
& \cellcolor{gray!25}52.94
& \cellcolor{gray!25}74.51
& \cellcolor{gray!25}74.16
& \cellcolor{gray!25}64.11
& \cellcolor{gray!25}67.53
& \cellcolor{gray!25}29.88
& \cellcolor{gray!25}67.78
& \cellcolor{gray!25}57.74 \\

& \cellcolor{gray!25}\soloq-N
& \cellcolor{gray!25}45.33
& \cellcolor{gray!25}45.22
& \cellcolor{gray!25}53.13
& \cellcolor{gray!25}74.59
& \cellcolor{gray!25}74.48
& \cellcolor{gray!25}64.55
& \cellcolor{gray!25}67.98
& \cellcolor{gray!25}25.61
& \cellcolor{gray!25}67.48
& \cellcolor{gray!25}57.60 \\

& \cellcolor{gray!25}\soloq-I
& \cellcolor{gray!25}45.56
& \cellcolor{gray!25}44.20
& \cellcolor{gray!25}52.21
& \cellcolor{gray!25}73.32
& \cellcolor{gray!25}73.94
& \cellcolor{gray!25}63.77
& \cellcolor{gray!25}67.24
& \cellcolor{gray!25}26.22
& \cellcolor{gray!25}66.34
& \cellcolor{gray!25}56.98 \\

\midrule


\multirow{10}{*}{LLaDA-1.5}
& FP16
& 47.20 & 87.46 & 75.02 & 72.93 & 74.86 & 65.47 & 66.05 & 50.61 & 83.47 & 69.23 \\
\cdashline{2-12}

& RTN
& 39.51 & 81.77 & 56.82 & 65.27 & 66.91 & 48.88 & 58.96 & 23.44 & 36.56 & 53.12 \\

& AWQ
& 40.96 & 82.22 & 66.84 & 67.93 & 69.32 & 51.12 & 60.03 & 30.07 & 57.95 & 58.49 \\

& QuaRot
& 43.21 & 84.23 & 65.34 & 69.55 & 70.17 & 56.23 & 57.66 & 37.33 & 65.86 & 61.06 \\

& DLLMQuant+
& 42.14 & 83.38 & 70.09 & 68.69 & 70.22 & 51.63 & 61.28 & 32.44 & 59.55 & 59.94 \\

& DLLMQuant++
& 43.87 & 84.18 & 69.20 & 71.74 & 73.58 & 57.27 & 60.04 & 44.58 & 74.33 & 64.31 \\

& STaR-Quant
& 48.77 & 85.12 & 72.43 & 72.09 & 73.36 & 63.29 & 64.34 & 46.53 & 76.45 & 66.93 \\

\cdashline{2-12}

& \cellcolor{gray!25}\soloq-C
& \cellcolor{gray!25}48.10
& \cellcolor{gray!25}86.44
& \cellcolor{gray!25}73.01
& \cellcolor{gray!25}71.59
& \cellcolor{gray!25}74.65
& \cellcolor{gray!25}64.60
& \cellcolor{gray!25}64.71
& \cellcolor{gray!25}46.95
& \cellcolor{gray!25}80.29
& \cellcolor{gray!25}67.82 \\

& \cellcolor{gray!25}\soloq-N
& \cellcolor{gray!25}48.09
& \cellcolor{gray!25}86.10
& \cellcolor{gray!25}73.86
& \cellcolor{gray!25}71.67
& \cellcolor{gray!25}74.76
& \cellcolor{gray!25}64.70
& \cellcolor{gray!25}65.38
& \cellcolor{gray!25}50.00
& \cellcolor{gray!25}81.80
& \cellcolor{gray!25}68.48 \\

& \cellcolor{gray!25}\soloq-I
& \cellcolor{gray!25}48.68
& \cellcolor{gray!25}87.46
& \cellcolor{gray!25}72.43
& \cellcolor{gray!25}71.27
& \cellcolor{gray!25}74.65
& \cellcolor{gray!25}64.12
& \cellcolor{gray!25}65.23
& \cellcolor{gray!25}43.90
& \cellcolor{gray!25}79.23
& \cellcolor{gray!25}67.44 \\

\bottomrule
\end{tabular}%
}
\end{table*}

\begin{table*}[h]
\caption{
Comparison with uniform INT4 quantization on Fast-dLLM v2.
\soloq-C, \soloq-N, and \soloq-I denote the codebook, NVFP4, and uniform INT4 variants, respectively.
All \soloq variants use the same unfolded K-RPBH topology.
Values are task accuracy (\%) and their average.
N/A indicates that the method is not applicable to the architecture.
}
\vspace{-2mm}
\label{tab:int4_fastdllm}
\small
\centering
\resizebox{\textwidth}{!}{%
\begin{tabular}{clcccccccccc}
\toprule
Model & Method
& Human-B
& Human-P
& MBPP-B
& MBPP-P
& GSM8K
& MATH
& IFEval
& MMLU
& GPQA
& Avg. \\
\midrule

\multirow{7}{*}{Fast-dLLM v2}
& FP16
& 65.85 & 59.76 & 61.90 & 52.65 & 84.15 & 60.30 & 61.18 & 66.54 & 30.80 & 60.35 \\
\cdashline{2-12}

& RTN
& 0.00 & 0.00 & 0.00 & 0.00 & 0.23 & 0.00 & 7.58 & 24.73 & 27.46 & 6.67 \\

& AWQ
& 0.00 & 0.00 & 0.00 & 0.00 & 0.00 & 0.00 & 9.06 & 24.98 & 25.45 & 6.61 \\

& QuaRot
& N/A & N/A & N/A & N/A & N/A & N/A & N/A & N/A & N/A & N/A \\

\cdashline{2-12}

& \cellcolor{gray!25}\soloq-C
& \cellcolor{gray!25}58.54
& \cellcolor{gray!25}53.05
& \cellcolor{gray!25}60.05
& \cellcolor{gray!25}50.53
& \cellcolor{gray!25}80.74
& \cellcolor{gray!25}54.86
& \cellcolor{gray!25}59.89
& \cellcolor{gray!25}64.73
& \cellcolor{gray!25}30.58
& \cellcolor{gray!25}57.00 \\

& \cellcolor{gray!25}\soloq-N
& \cellcolor{gray!25}59.10
& \cellcolor{gray!25}56.10
& \cellcolor{gray!25}57.40
& \cellcolor{gray!25}48.90
& \cellcolor{gray!25}82.64
& \cellcolor{gray!25}54.42
& \cellcolor{gray!25}55.27
& \cellcolor{gray!25}63.43
& \cellcolor{gray!25}33.71
& \cellcolor{gray!25}56.77 \\

& \cellcolor{gray!25}\soloq-I
& \cellcolor{gray!25}60.37
& \cellcolor{gray!25}56.10
& \cellcolor{gray!25}54.50
& \cellcolor{gray!25}48.94
& \cellcolor{gray!25}79.30
& \cellcolor{gray!25}52.74
& \cellcolor{gray!25}54.16
& \cellcolor{gray!25}61.41
& \cellcolor{gray!25}30.36
& \cellcolor{gray!25}55.32 \\

\bottomrule
\end{tabular}%
}
\vspace{-4mm}
\end{table*}

\newpage
\subsection{Folded Rotation for Efficient Deployment}
\label{apdix:fold}
\vspace{-2mm}
Online activation rotation introduces additional inference overhead, motivating us to examine whether the residual-stream rotation can be folded into model weights offline.
This is difficult in image or video diffusion transformers such as those targeted by OrbitQuant~\citep{lee2026orbitquant}, which commonly employ adaptive normalization such as AdaLN~\citep{peebles2023scalable}.
Let $x\in\mathbb{R}^{d}$ be a hidden state, $W\in\mathbb{R}^{m\times d}$ a linear weight, and
$\Pi\in\mathbb{R}^{d\times d}$ an orthogonal K-RPBH rotation.
We use $\mathcal{N}(\cdot)$ to denote RMS normalization without its learned channel-wise gain.
AdaLN can be written as
\begin{equation}
\mathrm{AdaLN}(x;c)
=
\mathrm{diag}\!\left(\gamma(c)\right)\mathcal{N}(x)+\beta(c),
\label{eq:adaln_fold}
\end{equation}
where the scale $\gamma(c)$ and shift $\beta(c)$ depend on the diffusion condition $c$.
Because these parameters vary at runtime, the residual-stream rotation cannot generally be absorbed into fixed adjacent weights and remains online.
In contrast, the dLLMs considered in \soloq use a static RMSNorm gain $\gamma$.
Since orthogonal rotations preserve the RMS norm, the full-precision operation can be equivalently written as
\begin{equation}
W\,\mathrm{diag}(\gamma)\mathcal{N}(x)
=
\underbrace{W\,\mathrm{diag}(\gamma)\Pi^\top}_{W_{\mathrm{fold}}}
\mathcal{N}(\Pi x).
\label{eq:fold_equiv}
\end{equation}
Thus, the RMSNorm gain and residual-stream rotation can be folded into $W_{\mathrm{fold}}$ offline, following the QuaRot-style topology and reducing online rotation overhead.
However, this equivalence does not generally survive quantization:
\begin{equation}
Q_W(W\Pi^\top)
Q_A\!\left(\Pi\,\mathrm{diag}(\gamma)\mathcal{N}(x)\right)
\neq
Q_W(W_{\mathrm{fold}})
Q_A\!\left(\mathcal{N}(\Pi x)\right).
\label{eq:fold_quant_mismatch}
\end{equation}
Tables~\ref{tab:fold_full_dllm} and~\ref{tab:fold_fastdllm} quantify this trade-off.
Although folding reduces online rotation cost, it introduces model-dependent accuracy degradation, with a relatively modest drop on LLaDA and a substantially larger drop on Fast-dLLM v2.
We therefore view folding as a deployment-time efficiency--accuracy trade-off rather than an accuracy-preserving transformation under low-bit quantization.

\begin{table*}[h]
\caption{
Comparison of unfolded and folded \soloq under W4A4 quantization on full-sequence diffusion language models.
\emph{Unfold} explicitly applies the activation-side rotations, whereas
\emph{Fold} absorbs foldable residual-stream rotations into adjacent weights offline.
}
\vspace{-2mm}
\label{tab:fold_full_dllm}
\small
\centering
\resizebox{\textwidth}{!}{%
\begin{tabular}{cllcccccccccc}
\toprule
Model & Method & Topology
& Truth.
& ARC-C
& Hella.
& Wino.
& PIQA
& MMLU
& C-EVAL
& Human.
& GSM8K
& Avg. \\
\midrule


\multirow{11}{*}{LLaDA-Base}
& FP16 & --
& 47.45 & 44.03 & 54.09 & 74.82 & 74.81 & 64.15 & 69.99 & 31.71 & 69.52 & 58.95 \\
\cdashline{2-13}

& RTN & --
& 40.45 & 41.83 & 45.40 & 64.72 & 67.95 & 49.26 & 57.95 & 14.02 & 16.56 & 44.23 \\

& AWQ & --
& 40.87 & 42.92 & 46.14 & 66.88 & 69.43 & 51.22 & 58.43 & 20.10 & 36.88 & 48.09 \\

& QuaRot & --
& 42.53 & 44.20 & 49.76 & 69.85 & 70.75 & 55.96 & 56.32 & 25.33 & 44.57 & 51.03 \\


& DLLMQuant++ & --
& 43.53 & 44.18 & 51.00 & 71.85 & 73.94 & 57.77 & 61.22 & 28.92 & 56.25 & 54.29 \\

& STaR-Quant & --
& 49.12 & 44.23 & 52.75 & 72.92 & 73.85 & 62.95 & 64.56 & 35.98 & 57.29 & 57.07 \\

\cdashline{2-13}

& \cellcolor{gray!25}\soloq-C
& \cellcolor{gray!25}Unfold
& \cellcolor{gray!25}45.22
& \cellcolor{gray!25}43.52
& \cellcolor{gray!25}52.94
& \cellcolor{gray!25}74.51
& \cellcolor{gray!25}74.16
& \cellcolor{gray!25}64.11
& \cellcolor{gray!25}67.53
& \cellcolor{gray!25}29.88
& \cellcolor{gray!25}67.78
& \cellcolor{gray!25}57.74 \\

& \cellcolor{gray!25}\soloq-C
& \cellcolor{gray!25}Fold
& \cellcolor{gray!25}43.72
& \cellcolor{gray!25}43.17
& \cellcolor{gray!25}51.00
& \cellcolor{gray!25}71.98
& \cellcolor{gray!25}73.45
& \cellcolor{gray!25}62.27
& \cellcolor{gray!25}66.05
& \cellcolor{gray!25}27.44
& \cellcolor{gray!25}64.06
& \cellcolor{gray!25}55.96 \\

& \cellcolor{gray!25}\soloq-N
& \cellcolor{gray!25}Unfold
& \cellcolor{gray!25}45.33
& \cellcolor{gray!25}45.22
& \cellcolor{gray!25}53.13
& \cellcolor{gray!25}74.59
& \cellcolor{gray!25}74.48
& \cellcolor{gray!25}64.55
& \cellcolor{gray!25}67.98
& \cellcolor{gray!25}25.61
& \cellcolor{gray!25}67.48
& \cellcolor{gray!25}57.60 \\

& \cellcolor{gray!25}\soloq-N
& \cellcolor{gray!25}Fold
& \cellcolor{gray!25}41.60
& \cellcolor{gray!25}43.52
& \cellcolor{gray!25}51.10
& \cellcolor{gray!25}72.30
& \cellcolor{gray!25}73.56
& \cellcolor{gray!25}61.84
& \cellcolor{gray!25}66.72
& \cellcolor{gray!25}24.39
& \cellcolor{gray!25}62.70
& \cellcolor{gray!25}55.30 \\

\midrule


\multirow{11}{*}{LLaDA-1.5}
& FP16 & --
& 47.20 & 87.46 & 75.02 & 72.93 & 74.86 & 65.47 & 66.05 & 50.61 & 83.47 & 69.23 \\
\cdashline{2-13}

& RTN & --
& 39.51 & 81.77 & 56.82 & 65.27 & 66.91 & 48.88 & 58.96 & 23.44 & 36.56 & 53.12 \\

& AWQ & --
& 40.96 & 82.22 & 66.84 & 67.93 & 69.32 & 51.12 & 60.03 & 30.07 & 57.95 & 58.49 \\

& QuaRot & --
& 43.21 & 84.23 & 65.34 & 69.55 & 70.17 & 56.23 & 57.66 & 37.33 & 65.86 & 61.06 \\


& DLLMQuant++ & --
& 43.87 & 84.18 & 69.20 & 71.74 & 73.58 & 57.27 & 60.04 & 44.58 & 74.33 & 64.31 \\

& STaR-Quant & --
& 48.77 & 85.12 & 72.43 & 72.09 & 73.36 & 63.29 & 64.34 & 46.53 & 76.45 & 66.93 \\

\cdashline{2-13}

& \cellcolor{gray!25}\soloq-C
& \cellcolor{gray!25}Unfold
& \cellcolor{gray!25}48.10
& \cellcolor{gray!25}86.44
& \cellcolor{gray!25}73.01
& \cellcolor{gray!25}71.59
& \cellcolor{gray!25}74.65
& \cellcolor{gray!25}64.60
& \cellcolor{gray!25}64.71
& \cellcolor{gray!25}46.95
& \cellcolor{gray!25}80.29
& \cellcolor{gray!25}67.82 \\

& \cellcolor{gray!25}\soloq-C
& \cellcolor{gray!25}Fold
& \cellcolor{gray!25}48.58
& \cellcolor{gray!25}86.78
& \cellcolor{gray!25}71.80
& \cellcolor{gray!25}70.56
& \cellcolor{gray!25}74.65
& \cellcolor{gray!25}63.47
& \cellcolor{gray!25}65.53
& \cellcolor{gray!25}46.34
& \cellcolor{gray!25}76.95
& \cellcolor{gray!25}67.18 \\

& \cellcolor{gray!25}\soloq-N
& \cellcolor{gray!25}Unfold
& \cellcolor{gray!25}48.09
& \cellcolor{gray!25}86.10
& \cellcolor{gray!25}73.86
& \cellcolor{gray!25}71.67
& \cellcolor{gray!25}74.76
& \cellcolor{gray!25}64.70
& \cellcolor{gray!25}65.38
& \cellcolor{gray!25}50.00
& \cellcolor{gray!25}81.80
& \cellcolor{gray!25}68.48 \\

& \cellcolor{gray!25}\soloq-N
& \cellcolor{gray!25}Fold
& \cellcolor{gray!25}49.16
& \cellcolor{gray!25}85.76
& \cellcolor{gray!25}72.11
& \cellcolor{gray!25}72.53
& \cellcolor{gray!25}74.97
& \cellcolor{gray!25}63.75
& \cellcolor{gray!25}65.23
& \cellcolor{gray!25}46.95
& \cellcolor{gray!25}79.23
& \cellcolor{gray!25}67.74 \\

\bottomrule
\end{tabular}%
}
\end{table*}

\begin{table*}[h]
\caption{
Comparison of unfolded and folded \soloq under W4A4 quantization on Fast-dLLM v2.
\emph{Unfold} explicitly applies activation-side rotations, whereas
\emph{Fold} absorbs foldable residual-stream rotations into adjacent weights offline.
Values are task accuracy (\%) and their average.
N/A indicates that the method is not applicable to the corresponding architecture.
}\vspace{-2mm}
\label{tab:fold_fastdllm}
\small
\centering
\resizebox{\textwidth}{!}{%
\begin{tabular}{cllcccccccccc}
\toprule
Model & Method & Topology
& Human-B
& Human-P
& MBPP-B
& MBPP-P
& GSM8K
& MATH
& IFEval
& MMLU
& GPQA
& Avg. \\
\midrule

\multirow{8}{*}{Fast-dLLM v2}

& FP16 & --
& 65.85
& 59.76
& 61.90
& 52.65
& 84.15
& 60.30
& 61.18
& 66.54
& 30.80
& 60.35 \\
\cdashline{2-13}

& RTN & --
& 0.00
& 0.00
& 0.00
& 0.00
& 0.23
& 0.00
& 7.58
& 24.73
& 27.46
& 6.67 \\

& AWQ & --
& 0.00
& 0.00
& 0.00
& 0.00
& 0.00
& 0.00
& 9.06
& 24.98
& 25.45
& 6.61 \\

& QuaRot & --
& N/A
& N/A
& N/A
& N/A
& N/A
& N/A
& N/A
& N/A
& N/A
& N/A \\

\cdashline{2-13}

& \cellcolor{gray!25}\soloq-C
& \cellcolor{gray!25}Unfold
& \cellcolor{gray!25}58.54
& \cellcolor{gray!25}53.05
& \cellcolor{gray!25}60.05
& \cellcolor{gray!25}50.53
& \cellcolor{gray!25}80.74
& \cellcolor{gray!25}54.86
& \cellcolor{gray!25}59.89
& \cellcolor{gray!25}64.73
& \cellcolor{gray!25}30.58
& \cellcolor{gray!25}57.00 \\

& \cellcolor{gray!25}\soloq-C
& \cellcolor{gray!25}Fold
& \cellcolor{gray!25}51.22
& \cellcolor{gray!25}47.56
& \cellcolor{gray!25}52.38
& \cellcolor{gray!25}44.97
& \cellcolor{gray!25}77.56
& \cellcolor{gray!25}49.38
& \cellcolor{gray!25}51.20
& \cellcolor{gray!25}56.70
& \cellcolor{gray!25}27.46
& \cellcolor{gray!25}50.94 \\

& \cellcolor{gray!25}\soloq-N
& \cellcolor{gray!25}Unfold
& \cellcolor{gray!25}59.10
& \cellcolor{gray!25}56.10
& \cellcolor{gray!25}57.40
& \cellcolor{gray!25}48.90
& \cellcolor{gray!25}82.64
& \cellcolor{gray!25}54.42
& \cellcolor{gray!25}55.27
& \cellcolor{gray!25}63.43
& \cellcolor{gray!25}33.71
& \cellcolor{gray!25}56.77 \\

& \cellcolor{gray!25}\soloq-N
& \cellcolor{gray!25}Fold
& \cellcolor{gray!25}59.10
& \cellcolor{gray!25}54.90
& \cellcolor{gray!25}52.40
& \cellcolor{gray!25}47.10
& \cellcolor{gray!25}74.60
& \cellcolor{gray!25}48.36
& \cellcolor{gray!25}50.65
& \cellcolor{gray!25}57.80
& \cellcolor{gray!25}30.36
& \cellcolor{gray!25}52.81 \\

\bottomrule
\end{tabular}%
}
\end{table*}

\FloatBarrier
\subsection{Attention sink in dLLM}
\label{sink}

\begin{figure}[h]
\centering
\includegraphics[width=0.9\linewidth]{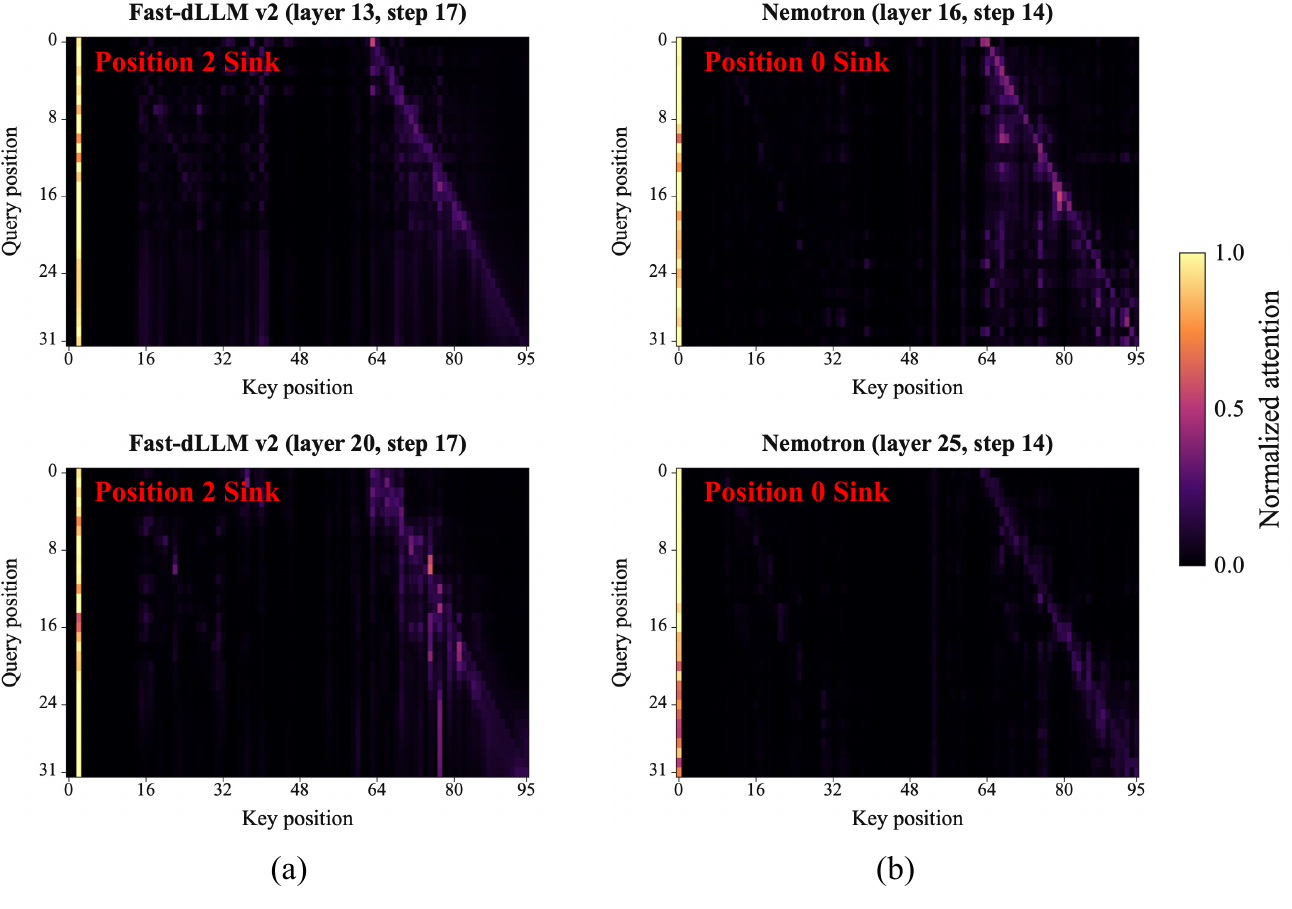}
\caption{
Attention-sink patterns in block-diffusion language models.
(a) Fast-dLLM v2 exhibits a concentrated sink at position 2 across representative layers.
(b) Nemotron-Labs-Diffusion exhibits a sink at position 0.
Brighter colors indicate larger attention weights.
}
\label{fig:sink}
\vspace{-2mm}
\end{figure}

Attention sinks, where a small number of token positions receive
disproportionately large attention mass, have also been observed in diffusion
language models~\citep{rulli2026attention}. Figure~\ref{fig:sink} shows that this behavior
persists in block-diffusion models: Fast-dLLM v2 exhibits a pronounced sink at
position 2, while Nemotron-Labs-Diffusion concentrates attention at position 0.
Because such positions can play a disproportionate role in attention
computation, aggressively quantizing their KV states may cause larger
degradation than quantizing ordinary positions.

Table~\ref{tab:sink_preservation} shows that preserving the attention-sink
position in full precision has little effect under W4A4KV4, suggesting that
4-bit KV quantization is sufficiently robust to the concentrated attention
mass. In contrast, under the more aggressive W4A4KV2 setting, preserving the
sink becomes beneficial. This indicates that sink-aware protection is mainly
important at ultra-low KV precision, while standard 4-bit KV quantization can
generally quantize the sink together with the remaining cache without
noticeable degradation.

\begin{table}[h]
\centering
\small
\caption{
Effect of preserving attention-sink KV states in full precision under
aggressive KV-cache quantization. Sink FP indicates that the identified
attention-sink position is kept in full precision while the remaining KV
states are quantized.
}
\label{tab:sink_preservation}
\begin{tabular}{lllcc}
\toprule
Model & Precision & Sink FP & GSM8K & IFEval \\
\midrule

\multirow{4}{*}{Fast-dLLM v2}
& \multirow{2}{*}{W4A4KV4}
& \cmark
& 80.59 & 56.75 \\
& 
& \xmark   & 79.98 & 56.38 \\
\cmidrule(lr){2-5}
& \multirow{2}{*}{W4A4KV2}
& \cmark & 55.80 & 41.22 \\
&
& \xmark   & 50.57 & 39.93 \\

\midrule

\multirow{4}{*}{Nemotron-Labs-Diffusion}
& \multirow{2}{*}{W4A4KV4}
& \cmark & 90.98 & 65.48 \\
&
& \xmark   & 91.28    & 65.06 \\
\cmidrule(lr){2-5}
& \multirow{2}{*}{W4A4KV2}
& \cmark & 82.94 & 54.18 \\
&
& \xmark   & 72.93 & 52.64 \\

\bottomrule
\end{tabular}
\vspace{-3mm}
\end{table}

\FloatBarrier
\vspace{-3mm}
\section{FPGA Implementation}
\label{fpga}
\vspace{-3mm}
\begin{figure}[h]
    \centering
    \includegraphics[width=\linewidth]{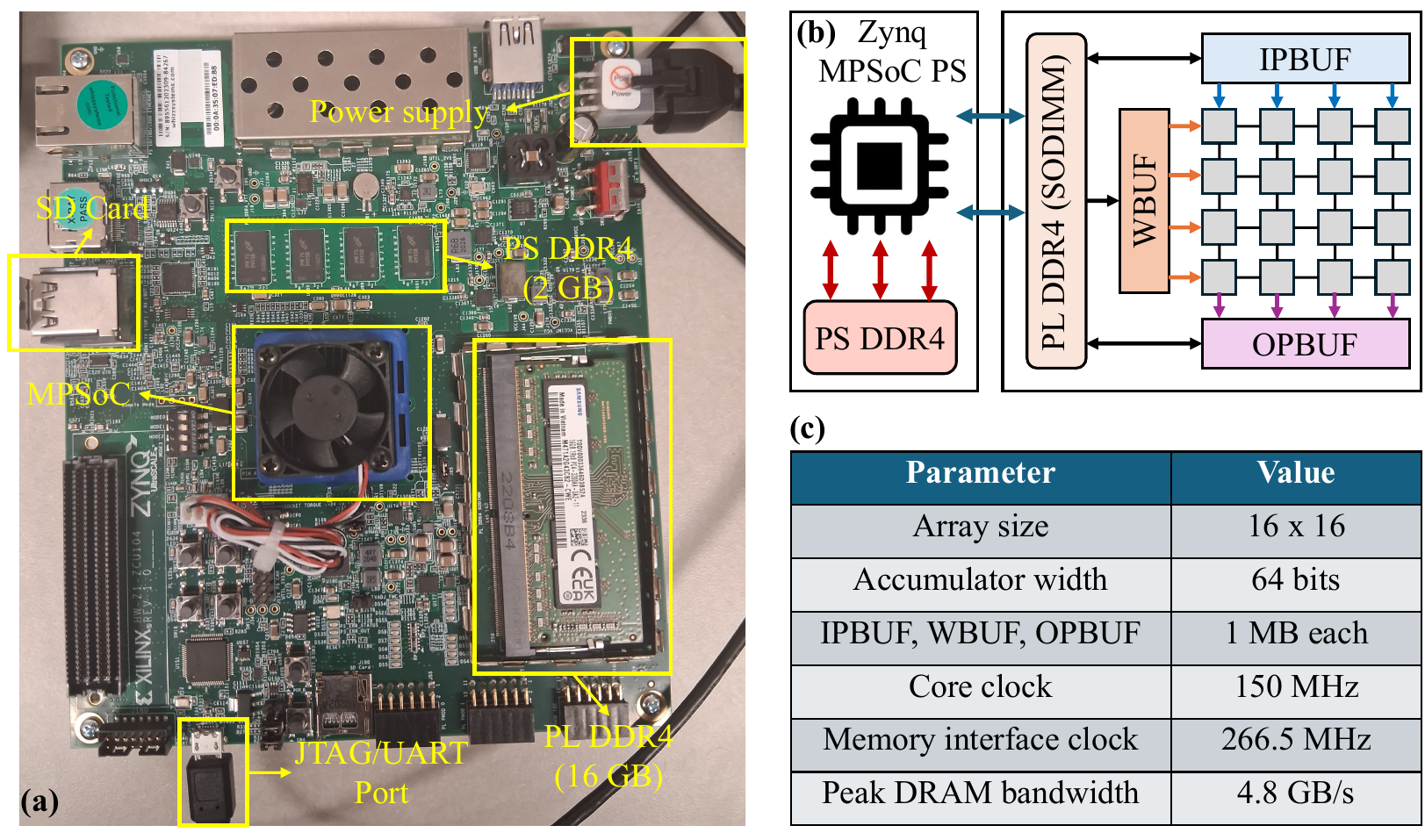}
    \caption{(a) Hardware setup on ZynQ Ultrascale+ ZCU104 FPGA board. (b) Micro-architecture of the hardware accelerator. (c) Hardware accelerator design parameters.}
    \label{fig:hw_setup}
\end{figure}

\begin{figure}[h]
\centering
\begin{minipage}[c]{0.48\linewidth}
\centering
\includegraphics[width=\linewidth]{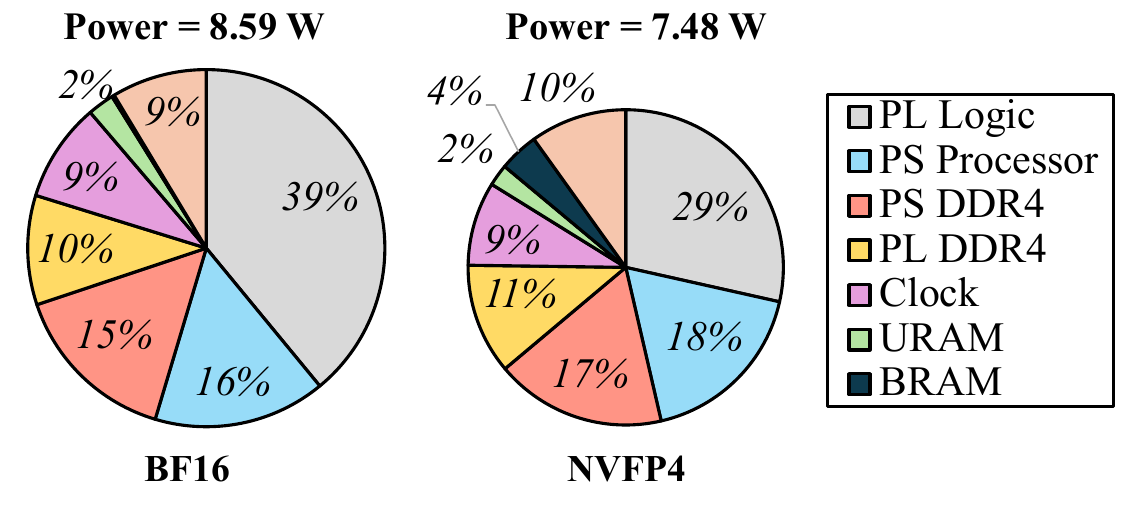}
\caption{Power breakdown of BF16 and NVFP4 accelerators for LLaDA-Base.}
\label{fig:power_soloq}
\end{minipage}%
\hfill
\begin{minipage}[c]{0.48\linewidth}
\centering
\captionof{table}{Throughput and energy efficiency of BF16 and NVFP4 accelerators on LLaDA-Base.}
\label{tab:throughput_energy}
\begin{tabular}{lrr}
\toprule
\textbf{Metric} & \textbf{BF16} & \textbf{NVFP4} \\
\midrule
Tokens/sec & 0.13 ($1\times$) & \textbf{\boldmath 0.252 ($1.94\times$)} \\
Tokens/kJ & 15.13 ($1\times$) & \textbf{\boldmath 40.90 ($2.70\times$)} \\
\bottomrule
\end{tabular}
\end{minipage}
\end{figure}
We implement a systolic-array accelerator with custom BF16 and NVFP4 datapaths
on a Zynq UltraScale+ ZCU104 FPGA and evaluate LLaDA-Base-8B for edge dLLM
inference. Fig.~\ref{fig:hw_setup}(a) shows the hardware setup. The Zynq MPSoC
consists of a processing system (PS) and programmable logic (PL), each with
access to external DDR4 memory. The model weights are distributed across the
PS and PL DDR4 memories, while activations are stored in the PL DDR4.

As shown in Fig.~\ref{fig:hw_setup}(b), the systolic-array accelerator is
implemented in the PL and receives execution instructions and a portion of the
model weights from the PS. The systolic array comprises multiply-and-accumulate
(MAC) units together with input (IP), weight (W), and output (OP) buffers to
perform tiled matrix multiplication on data tiles fetched from the PL DDR4.

Fig.~\ref{fig:hw_setup}(c) summarizes the design specifications of the hardware
accelerator. For the NVFP4 accelerator, we double the compute capacity because
NVFP4 enables higher compute density than BF16.
Fig.~\ref{fig:power_soloq} shows the power breakdown of the accelerators during
LLaDA-Base-8B inference, where the compute logic, DDR memories, PS processor,
and clock generation and routing are the major contributors. The NVFP4
accelerator consumes 15\% less power while providing twice the compute
throughput of the BF16 design. As shown in
Table~\ref{tab:throughput_energy}, \soloq's NVFP4 accelerator achieves a
$\mathbf{1.94\times}$ end-to-end speedup and
$\mathbf{2.70\times}$ higher energy efficiency than BF16 on LLaDA-Base-8B.

\lstdefinestyle{qualcode}{
    basicstyle=\ttfamily\scriptsize,
    columns=fullflexible,
    keepspaces=true,
    breaklines=true,
    frame=none,
    xleftmargin=0pt,
    xrightmargin=0pt,
    aboveskip=0pt,
    belowskip=0pt,
    showstringspaces=false
}

\section{Qualitative Generation Examples}
\label{app:qualitative}

We provide generations to illustrate how low-bit quantization
affects model outputs beyond aggregate benchmark scores. The examples cover
mathematical reasoning, code generation, and general instruction following,
including both agreement and divergent outputs among FP16, \soloq-C, and
\soloq-N. The mathematical reasoning examples are generated with Fast-dLLM
v2-7B, while the code-generation examples use LLaDA-8B-Base. All variants use
identical prompts and model-specific deterministic decoding configurations.
These examples are selected for qualitative inspection only. All quantitative
claims in the paper are based on the complete benchmark results.


\noindent\rule{\linewidth}{0.5pt}

\noindent
\textit{Mathematical reasoning --- Divergent outputs}

\vspace{0.7mm}

\noindent
\textbf{Prompt.}
\textit{
A robe takes 2 bolts of blue fiber and half that much white fiber.
How many bolts in total does it take?
}

\noindent\dotfill

\noindent
\textbf{FP16} \quad \cmark

\vspace{0.3mm}

White fiber $=2/2=1$ bolt. Total $=2+1=3$ bolts.
Final answer: $\boxed{3}$.

\noindent\dotfill

\noindent
\textbf{\soloq-C} \quad \cmark


$B=2$, $W=B/2=1$, and $B+W=2+1=3$.
Final answer: $\boxed{3}$.

\noindent\dotfill

\noindent
\textbf{\soloq-N} \quad \cmark

\vspace{0.3mm}

White fiber $=(1/2)\times2=1$ bolt. Total $=2+1=3$.
Final answer: $\boxed{3}$.

\vspace{1mm}


\noindent\rule{\linewidth}{0.5pt}

\noindent
\textit{Code generation --- Correct Outputs}

\vspace{0.7mm}

\noindent
\textbf{Prompt.}
\textit{
Implement \texttt{has\_close\_elements(numbers, threshold)}.
}

\noindent\dotfill


\noindent
\textbf{FP16} \quad \cmark
\vspace{0.3mm}

\begin{lstlisting}[style=qualcode,language=Python]
if len(numbers) < 2:
    return False
for i in range(len(numbers)):
    for j in range(i + 1, len(numbers)):
        if abs(numbers[i] - numbers[j]) < threshold:
            return True
return False
\end{lstlisting}

\noindent\dotfill

\noindent
\textbf{\soloq-C} \quad \cmark
\vspace{0.3mm}

\begin{lstlisting}[style=qualcode,language=Python]
if len(numbers) < 2:
    return False
sorted_numbers = sorted(numbers)
for i in range(len(sorted_numbers) - 1):
    if abs(sorted_numbers[i] - sorted_numbers[i + 1]) < threshold:
        return True
return False
\end{lstlisting}

\noindent\dotfill

\noindent
\textbf{\soloq-N} \quad \cmark
\vspace{0.3mm}

\begin{lstlisting}[style=qualcode,language=Python]
if len(numbers) < 2:
    return False
for i in range(len(numbers)):
    for j in range(i + 1, len(numbers)):
        if abs(numbers[i] - numbers[j]) < threshold:
            return True
return False
\end{lstlisting}



\noindent\rule{\linewidth}{0.5pt}

\noindent
\textit{Mathematical reasoning --- Incorrect Outputs}


\noindent
\textbf{Prompt.}
\textit{
Josh decides to try flipping a house. He buys a house for \$80,000 and then puts in \$50,000 in repairs. This increased the value of the house by 150\%. How much profit did he make?
}

\noindent\dotfill

\noindent
\textbf{FP16} \quad \xmark


Computes $80{,}000+50{,}000=130{,}000$, then
$130{,}000-50{,}000=80{,}000$.
Final answer: $\boxed{80000}$.

\noindent\dotfill

\noindent
\textbf{\soloq-C} \quad \xmark


Computes the new value as
$80{,}000+1.5(80{,}000)=200{,}000$, then
$200{,}000-80{,}000-50{,}000=170{,}000$.
Final answer: $\boxed{170000}$.

\noindent\dotfill

\noindent
\textbf{\soloq-N} \quad \cmark


Computes total cost as $80{,}000+50{,}000=130{,}000$ and the new value as
$80{,}000+120{,}000=200{,}000$, giving a profit of
$200{,}000-130{,}000=70{,}000$.
Final answer: $\boxed{70000}$.

\vspace{1mm}

















\noindent\rule{\linewidth}{0.5pt}

\noindent
\textit{General explanation --- length-constrained instruction following}

\vspace{0.7mm}

\noindent
\textbf{Prompt.}
\textit{
Explain to a non-specialist how AI models can be made to use less memory and
respond faster. Describe a few common approaches, give a simple everyday
example, and explain the trade-offs between speed, cost, and answer quality.
Use clear language and keep your answer to about 250--300 words.
}

\noindent\dotfill


\noindent
\textbf{FP16}

\vspace{0.3mm}

{\scriptsize
Sure, I'd be happy to explain this in simple terms! AI models can be made to
use less memory and respond faster through several approaches. Here are a few:

\textbf{1. Model Compression}: This involves reducing the size of the model
without significantly impacting its performance. For example, you can remove
less important connections in a neural network, or use techniques like
quantization to reduce the precision of the weights. Imagine compress a JPEG
image - you can reduce its size without losing much quality. In AI, this can
help reduce memory usage and improve response times.

\textbf{2. Efficient Model Architect}: Some models are designed to be more
efficient from the. For instance, MobileNet is a convolutional neural network
designed for mobile devices. It uses fewer parameters and computations than
traditional models, making it faster and less memory-intensive.

\textbf{3. Hardware Optimization}: This involves using hardware that's
optimized for AI tasks, like GPUs or TPUs. These can perform complex
calculations much faster than regular CPUs, reducing response times.

\textbf{4. Batch Processing}: This involves processing multiple inputs at once,
rather than one by one. This can reduce the overhead of individual function
calls and improve memory usage.

The trade-offs between speed, cost, and answer quality are important. For
example, compressing a model can reduce memory usage and improve response
times, but it might also reduce the accuracy of the model. Using an efficient
model architecture can improve speed and cost, but it might require more
expertise to design and implement. Hardware optimization can improve speed and
cost, but it might require expensive hardware. Batch processing can improve
speed and cost, but it might require more memory.

In conclusion, there are many ways to make AI models use less memory and
respond faster, but each has its trade-offs. It's important to consider the
specific requirements and constraints of the application when deciding which
approach to use.

\textit{(312 words; 395 tokens; EOS)}
}

\noindent\dotfill


\noindent
\textbf{\soloq-C}


{\scriptsize
Sure, I'd be happy to explain this in simple terms!

AI models can be made to use less memory and respond faster through several
approaches. Here are a few:

\textbf{1. Model Compression}: This is like compress a big, complex book into
a smaller, simpler one. You can do this by removing unnecessary parts, using
shared layers, or applying techniques like pruning or quantization. For
example, a model that originally used 1GB of memory might be compressed to
100MB, making it run faster and use less memory.

\textbf{2. Efficient Architectures}: These are like using a specialized tool
for a specific job. For instance, MobileNet is designed for mobile devices
with limited resources. It uses fewer parameters and computations than other
models, making it faster and more efficient.

\textbf{3. Hardware-Aware Training}: This is like designing a racecar for a
specific track. By training the model on the target hardware, you can optimize
it for that hardware, improving its speed and efficiency.

There are some trade-offs to consider:

\textbf{Speed}: Fasteraller models generally run faster, but they might not be
as accurate as larger models. For example, a smaller image model might take
longer to train but could be faster during inference.

\textbf{Cost}: Smaller models can be cheaper to train and deploy, but they
might require more computational resources to achieve similar accuracy.

\textbf{Answer Quality}: Smaller models might provide less detailed or accurate
answers than larger models. For instance, a smaller language model might
struggle to understand complex sentences or generate nuanced responses.

In conclusion, the key is to balance the need for speed and memory efficiency
with the desire for accurate and detailed answers.

\textit{(274 words; 363 tokens; EOS)}
}

\noindent\dotfill


\noindent
\textbf{\soloq-N} 


{\scriptsize
Sure, I'd be happy to explain this in simple terms!

AI models can be made to use less memory and respond faster through several
approaches. Here are a few common ones:

\textbf{1. Model Compression}: This involves reducing the size of the model by
removing redundant weights or using more efficient representations. For
example, you can use a technique called ``pruning'' to remove less important
weights from the model, making it smaller and faster. Another technique is
``quantization'', where you reduce the precision of the weights, also making
the model smaller and faster.

\textbf{2. Knowledge Distillation}: This is a process where a smaller model is
trained to mimic the behavior of a larger model. For instance, you can train a
simple model to copy the predictions of a complex model, but in a way that the
simple model can make those predictions faster and with less memory.

\textbf{3. Hardware Optimization}: This involves optimizing the hardware to run
the model more efficiently. For example, you can use a specialized chip like a
GPU or a tensor processing unit (TPU) to speed the model up and reduce memory
usage.

Let's consider a simple everyday example. Imagine you have a recipe for a
complex dish that takes a long time to prepare and requires a lot of
ingredients. You can make it simpler (model compression) by removing some of
the ingredients (weights), make it smaller (knowledge distillation) by
teaching a simpler recipe to mimic the complex one, or optimize the
preparation process (hardware optimization) by using a more efficient kitchen.

However, there are trade-offs. Making the model simpler or smaller might
reduce the quality (answer quality) of the predictions. For example, a simpler
model might not capture as much complexity in the data, leading to less
accurate predictions. Similarly, making the model faster might require more
resources (cost) to run, or it might not be as accurate as a slower model.
Therefore, it's important to balance speed, cost, and answer quality when
optimizing AI models.

\textit{(331 words; 427 tokens; EOS)}
}

\vspace{1mm}
\noindent\rule{\linewidth}{0.5pt}

\vspace{1mm}

The examples illustrate that low-bit perturbations can affect both task-level
correctness and generation behavior. In the agreement cases, quantized models
preserve the reasoning or functional behavior of the full-precision model. In
the divergent cases, small representation perturbations alter discrete token
decisions, sometimes correcting an FP16 failure and sometimes introducing a
new error. The general instruction-following example further shows that the
overall structure and content of a response can remain similar across
precisions while details such as wording, examples, and response length vary.
These observations motivate evaluating quantization over complete benchmark
suites rather than relying on individual generations.

\section{Limitations}
\label{limit}
\soloq introduces online activation rotation to transform representations into the K-RPBH basis. Although weight rotations are performed offline, activation-side rotations incur additional inference overhead. Folding these rotations into adjacent weights can reduce this cost, but, as shown in Appendix~\ref{apdix:fold}, the folded topology may degrade accuracy under low-bit quantization, particularly for block-diffusion models.  
In addition, \soloq-C uses non-uniform codebooks whose centroids are represented losslessly in INT8 for efficient computation. Consequently, its low-bit representation cannot directly exploit native 4-bit Tensor Core execution and instead relies on 8-bit kernels, resulting in lower speedups than the hardware-native NVFP4 path of \soloq-N. Finally, the full latency benefit of \soloq-N depends on hardware support for native NVFP4 computation; on devices without such support, its practical acceleration may be reduced.

\end{document}